%% file: main.tex
\documentclass[10pt,twocolumn,letterpaper]{article}
\input{preamble}
\input{authors}
\title{Beyond Feature Fidelity: Recovery and Readout Adaptation in Multimodal Large Language Models}
\date{}
\begin{document}
\twocolumn[{
\vspace*{-20pt}
\noindent\makebox[\textwidth][s]{%
\raisebox{-0.5\height}{\includegraphics[height=57pt]{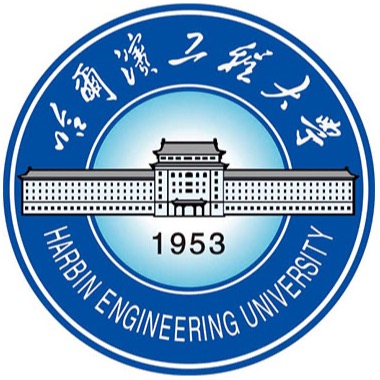}}\hfill
\raisebox{-0.5\height}{\includegraphics[height=54pt]{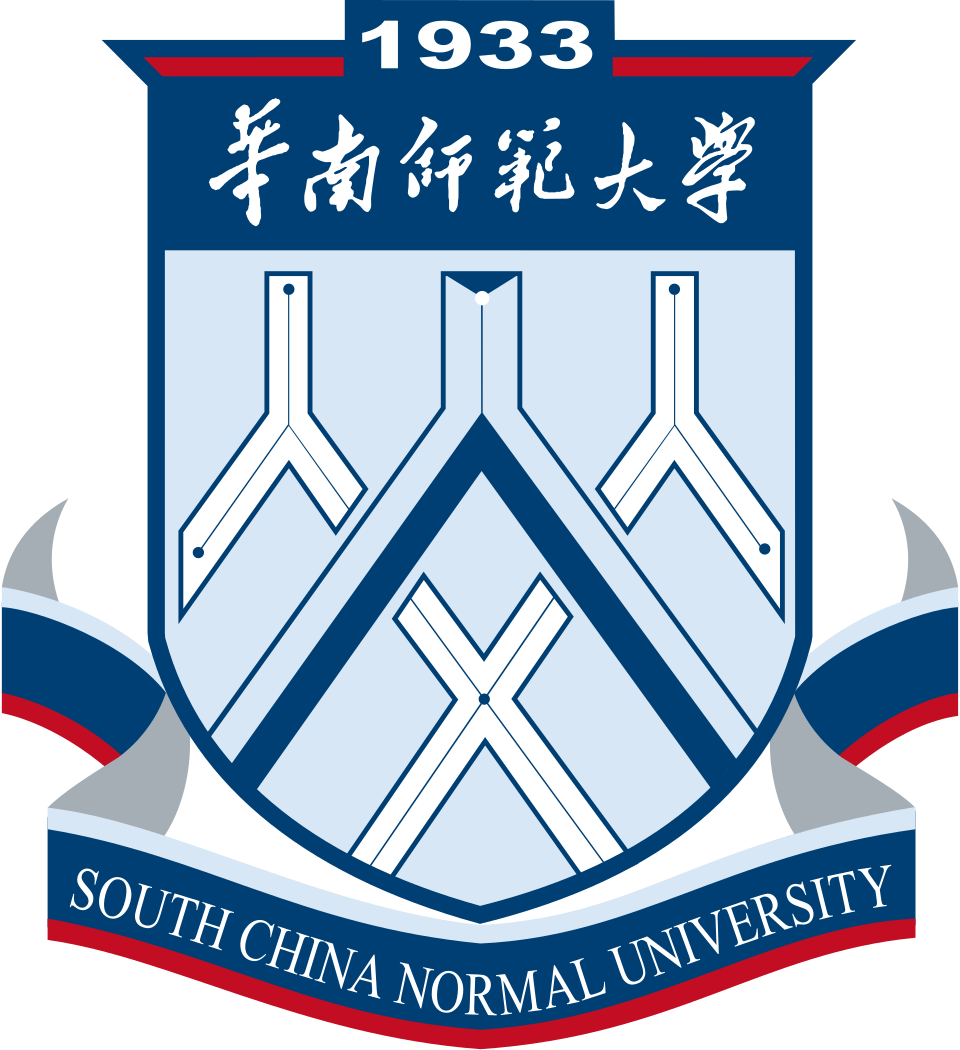}}\hfill
\raisebox{-0.5\height}{\includegraphics[height=51pt]{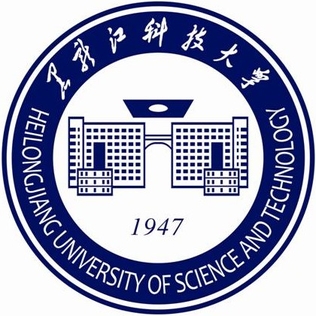}}\hfill
\raisebox{-0.5\height}{\includegraphics[height=50pt]{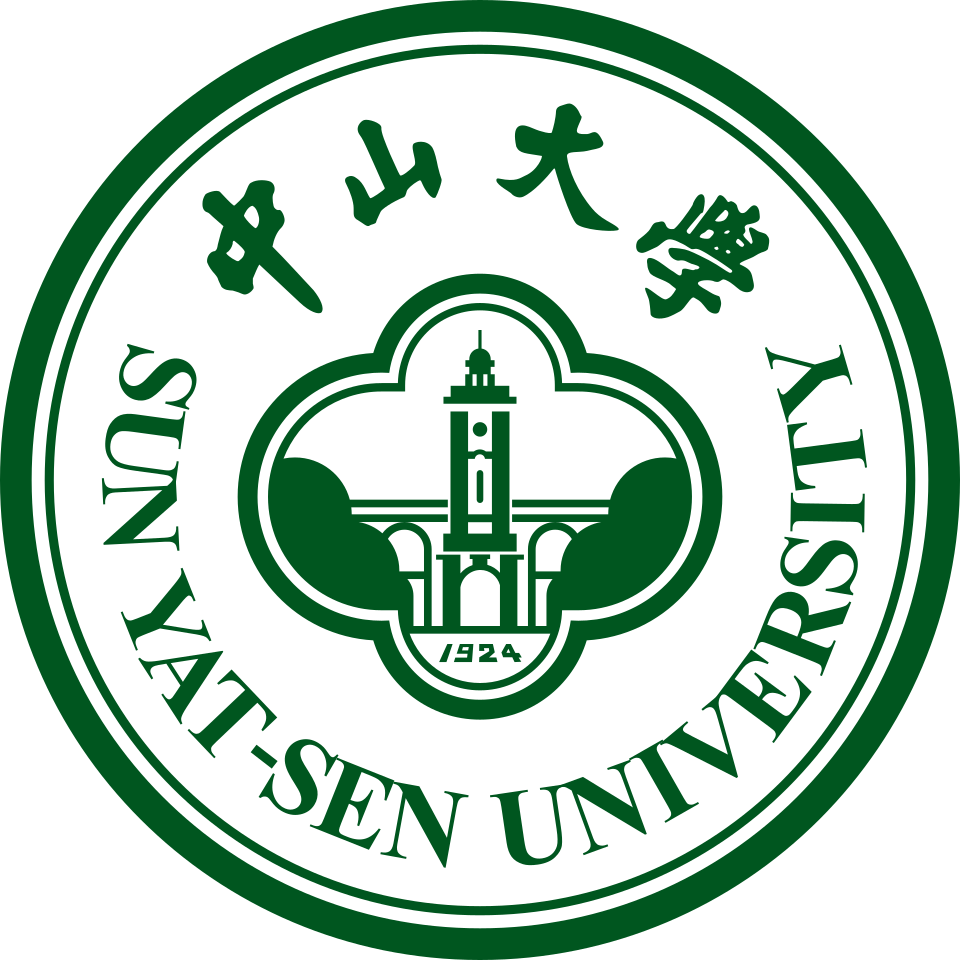}}\hfill
\raisebox{-0.5\height}{\includegraphics[height=48pt]{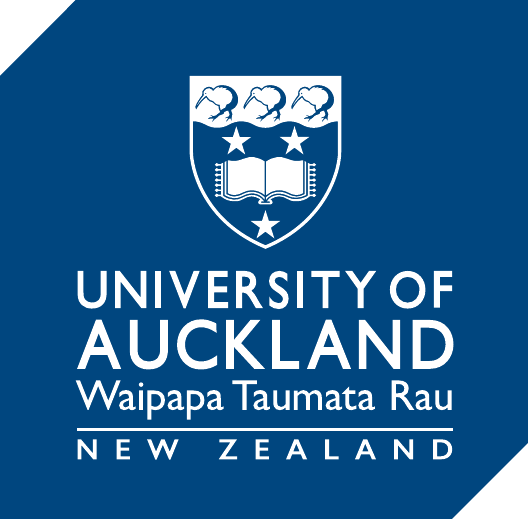}}\hfill
\raisebox{-0.5\height}{\includegraphics[height=36pt]{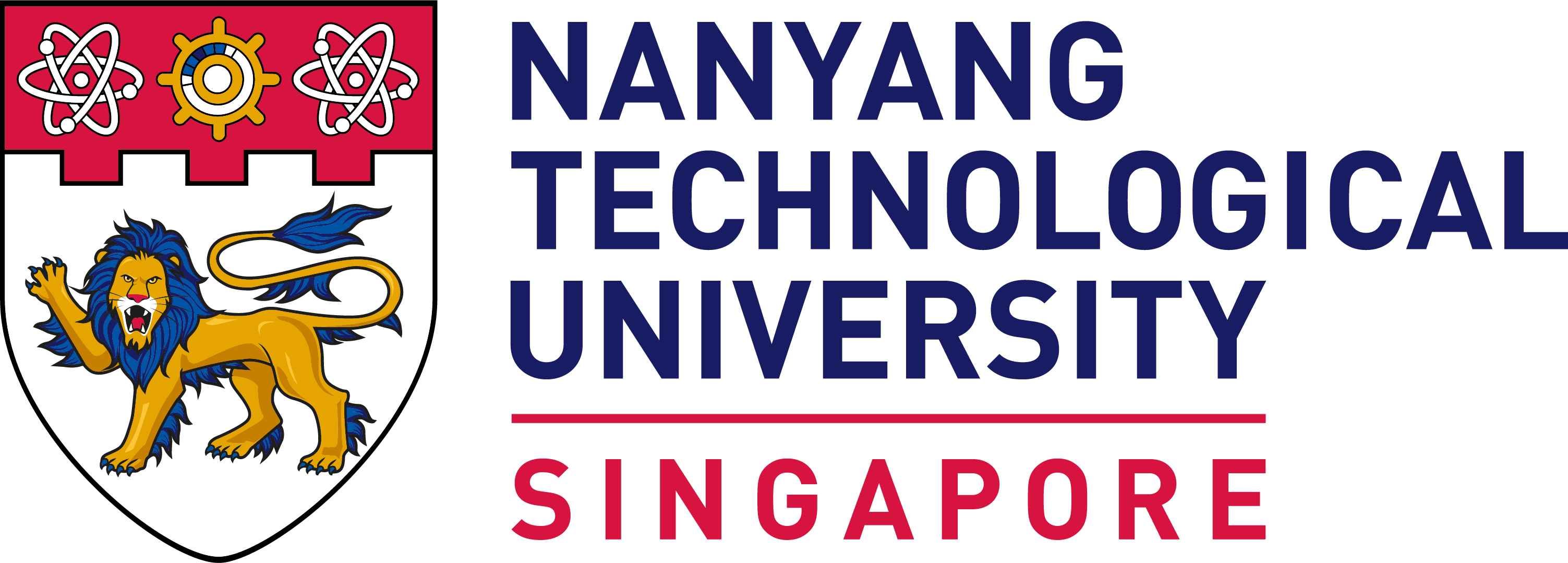}}%
}\par
\vspace{-0.7em} 
\begin{center}
{\color{preprintlink}\rule{\textwidth}{1.6pt}}\par
\vspace{0.8em}
{\fontsize{20}{24}\selectfont\bfseries\color{preprintlink}
Beyond Feature Fidelity:\\
Recovery and Readout Adaptation\\
in Multimodal Large Language Models\par}
\vspace{0.7em}
{\color{preprintlink}\rule{\textwidth}{0.45pt}}\par
\vspace{0.75em}
{\fontsize{10}{14}\selectfont\PreprintAuthors\par}
\vspace{0.65em}
{\fontsize{9}{12}\selectfont\PreprintAffiliations\par}
\vspace{0.4em}
{\footnotesize\PreprintAuthorNotes\par}
\vspace{0.35em}
{\footnotesize\href{mailto:zhuzhiliang@hrbeu.edu.cn}{\texttt{zhuzhiliang@hrbeu.edu.cn}}\quad\href{mailto:hjz839140159@gmail.com}{\texttt{hjz839140159@gmail.com}}\par}
\vspace{0.4em}
{\small\href{https://github.com/allenzg/beyond-feature-fidelity}{\textcolor{preprintlink}{\textbf{GitHub:}\;\texttt{github.com/allenzg/beyond-feature-fidelity}}}\par}
\end{center}
\vspace{0.35em}
\begin{minipage}{\textwidth}
\begin{center}{\large\bfseries\color{preprintlink}Abstract}\end{center}
\input{sec/0_abstract}
\end{minipage}
\vspace{1.6em}
}]
\input{sec/1_introduction}
\input{sec/2_related_work}
\input{sec/3_methodology}
\input{sec/4_experiments}
\input{sec/5_conclusion}
{\small\raggedright
\bibliographystyle{ieeenat_fullname}
\bibliography{references}
}
\clearpage
\onecolumn
\appendix
\begin{center}
{\Large\bfseries\color{preprintlink}Supplementary Material}\par
\vspace{0.5em}
{\color{preprintlink}\rule{\textwidth}{0.45pt}}
\end{center}
\let\appendixsection\section
\renewcommand{\section}{\FloatBarrier\appendixsection}
\let\appendixsubsection\subsection
\renewcommand{\subsection}{\FloatBarrier\appendixsubsection}
\input{sec/appendix}

\end{document}

%% file: preamble.tex
\usepackage{times}
\usepackage[T1]{fontenc}
\usepackage[letterpaper,top=0.8in,bottom=0.8in,left=0.75in,right=0.75in,columnsep=0.28in]{geometry}
\usepackage{amsmath,amssymb}
\usepackage{graphicx}
\usepackage{booktabs,multirow}
\usepackage[table,dvipsnames]{xcolor}
\usepackage{xspace}
\usepackage[numbers,sort&compress]{natbib}
\usepackage[font=small,labelfont=bf,labelsep=period]{caption}
\usepackage[hyphens]{url}
\definecolor{preprintlink}{RGB}{88,60,140}
\usepackage[colorlinks=true,allcolors=preprintlink,breaklinks=true]{hyperref}
\hypersetup{
    pdftitle={Beyond Feature Fidelity: Recovery and Readout Adaptation in Multimodal Large Language Models},
    pdfauthor={Zhiliang Zhu, Jiazheng Huang, Chubin Peng, Zijie Yang, Puyue Wang, Yuquan Xue, Dawei Zhao}
}
\makeatletter
\renewcommand\section{\@startsection{section}{1}{\z@}%
    {-2.5ex plus -0.6ex minus -0.2ex}{1.1ex plus 0.2ex}{\normalfont\large\bfseries\color{preprintlink}}}
\renewcommand\subsection{\@startsection{subsection}{2}{\z@}%
    {-2.0ex plus -0.5ex minus -0.2ex}{0.8ex plus 0.2ex}{\normalfont\normalsize\bfseries\color{preprintlink}}}
\renewcommand\paragraph{\@startsection{paragraph}{4}{\z@}%
    {0.8em}{-1em}{\normalfont\normalsize\bfseries}}
\makeatother

\usepackage{array,tabularx,subcaption,placeins,microtype}
\input{math_commands}

\definecolor{bestblue}{RGB}{198,218,239}
\definecolor{secondblue}{RGB}{225,237,248}
\newcommand{\best}[1]{\cellcolor{bestblue}\textbf{#1}}
\newcommand{\second}[1]{\cellcolor{secondblue}#1}

%% file: math_commands.tex
\usepackage{amsmath,amsfonts,bm}

\def\eqref#1{equation~\ref{#1}}

\def\1{\bm{1}}

\DeclareMathAlphabet{\mathsfit}{\encodingdefault}{\sfdefault}{m}{sl}
\SetMathAlphabet{\mathsfit}{bold}{\encodingdefault}{\sfdefault}{bx}{n}



%% file: authors.tex
\newcommand{\PreprintAuthors}{%
Zhiliang Zhu\textsuperscript{1}, Jiazheng Huang\textsuperscript{2}, Chubin Peng\textsuperscript{3}, Zijie Yang\textsuperscript{4},\\
Puyue Wang\textsuperscript{5}, Yuquan Xue\textsuperscript{6}, Dawei Zhao\textsuperscript{1,\textdagger}}
\newcommand{\PreprintAffiliations}{%
\textsuperscript{1}Harbin Engineering University\quad
\textsuperscript{2}South China Normal University\\
\textsuperscript{3}Heilongjiang University of Science and Technology\quad
\textsuperscript{4}Sun Yat-sen University\\
\textsuperscript{5}University of Auckland\quad
\textsuperscript{6}Nanyang Technological University}

\newcommand{\PreprintAuthorNotes}{%
\textsuperscript{\textdagger}Corresponding author.}

%% file: sec/0_abstract.tex
Feature recovery targets the geometry of a clean visual representation;
multimodal prediction depends on how a readout uses that representation.
We examine this distinction in controlled systems connecting a shared
visual encoder to frozen language-side models through learned readouts.
Across four downstream models and nine benchmarks, recovery has mixed
effects under a clean-trained readout. Readout adaptation improves the
reported utility in most evaluated model--benchmark combinations, yet
adapting directly to raw features is often competitive with or better
than adapting after explicit recovery. In a GQA control, additional
readout training on recovered clean features raises the score from
$58.82$ to $59.35$, while adaptation using recovered clean and corrupted
features reaches $59.67$. This limits attribution of the entire adaptation
gain to distribution matching. Separate fidelity analyses characterize
task-specific and task-agnostic recovery settings. Together, the results
motivate evaluating the representation--readout pair through distinct
measurements of feature fidelity, fixed-readout utility, and
adapted-readout utility.

%% file: sec/1_introduction.tex
\section{Introduction}

Many contemporary multimodal large language models (MLLMs) couple a visual encoder with a language model via a learned vision--language interface
\citep{liu2023visual,li2023blip2}. Under visual distribution shifts such as
blur, noise, compression, and exposure variation, the distribution of encoder
features can change \citep{chhipa2023self,tang2026robustu1}. Recovering these
features toward their clean counterparts is a natural robustness strategy, but
its usefulness also depends on whether the downstream interface can consume
the recovered representation.

Feature-recovery methods are typically optimized and evaluated with
representation-space criteria such as reconstruction error, cosine similarity,
and structural consistency \citep{yang2024dvt,parikh2026latent}. These metrics
measure \emph{representation fidelity}, rather than whether the recovered
information remains usable by a fixed multimodal interface. Thus, a recovered
representation can move closer to its clean counterpart without improving
downstream utility.

We study this interaction in a controlled DINOv3--LLaVA-1.5-7B
system \citep{simeoni2025dinov3,liu2024improved}, comparing raw
and recovered features under clean-trained and adapted readouts.
On GQA \citep{hudson2019gqa}, inserting recovery changes
Avg-All-C from $58.99$ to $58.82$, while adapting the readout
to recovered inputs raises the score to $59.67$. Direct
adaptation to raw features reaches $59.74$. These observations
motivate distinguishing three effects: recovery under a fixed
readout, adaptation to a fixed representation, and the additional
value of recovery when both readouts are adapted.

\begin{figure*}[t]
    \centering
    \includegraphics[width=\linewidth]{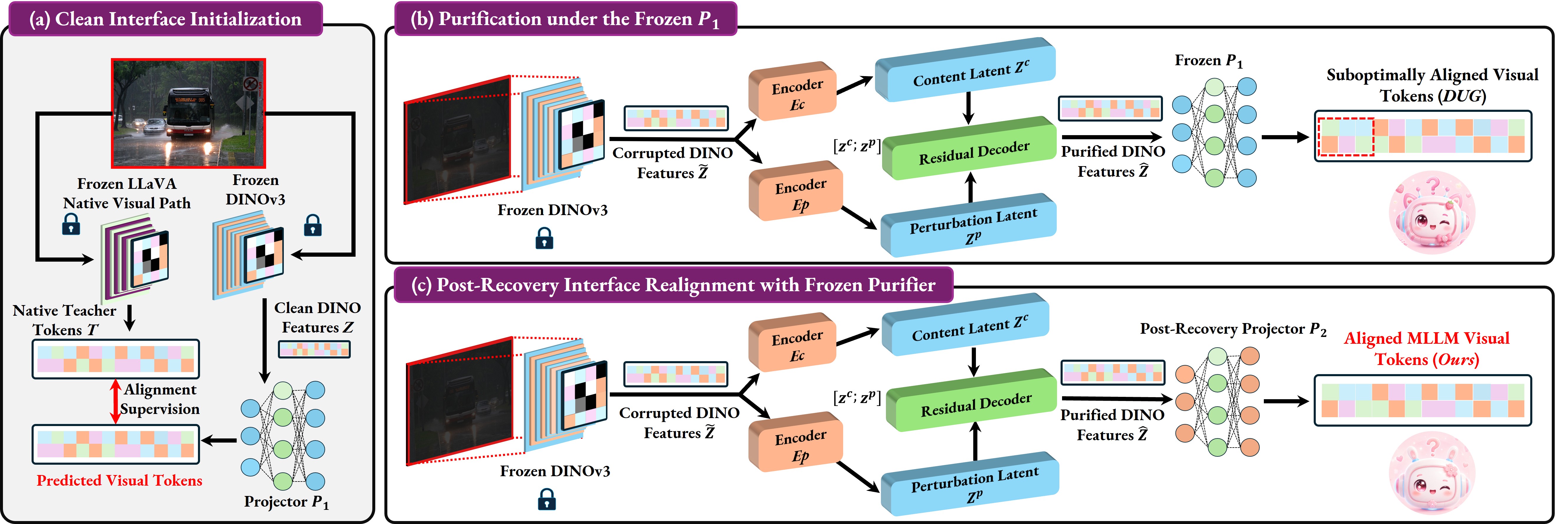}
    \caption{Controlled evaluation of representation recovery and readout adaptation in multimodal systems.}
    \label{fig:dug_overview}
\end{figure*}

Limited utility after recovery can arise from changes in task-relevant
information, readout sensitivity to the transformed features, or both.
We investigate these possibilities with a controlled $2\times2$ design
that varies the representation supplied to the readout and whether the
readout is adapted. Holding the recovered features fixed while updating
the readout measures the utility effect of that update. A clean-only
continuation control further tests how much of the observed gain can
also arise without including corrupted inputs in adaptation. These
comparisons establish readout dependence, while not uniquely separating
information loss from interface mismatch.

We evaluate these controlled effects across multiple MLLM families and
downstream tasks. Recovery under the clean-trained readout has heterogeneous
and often negative utility effects, whereas readout adaptation improves
utility in most evaluated model--benchmark combinations. Moreover,
adapting directly to raw corrupted features is frequently competitive with
adapting after explicit recovery, indicating that recovery and readout
adaptation contribute distinct effects. We further examine these observations
when recovery is trained on an unlabeled, task-agnostic subset of Open Images
\citep{kuznetsova2020openimages} and transferred without downstream question--answer supervision, as well as under different adaptation-data scales.

Our contributions are threefold:
\begin{itemize}
    \item We introduce a controlled representation--readout evaluation that distinguishes the utility effect of recovery under a fixed
    readout, the effect of readout adaptation, and the marginal benefit
    of recovery when both readouts are adapted.

    \item Across four downstream models and nine benchmarks, we show that recovery-only utility effects are heterogeneous, readout adaptation
    improves recovered-input utility in most evaluated settings, and explicit
    recovery provides no consistent advantage over direct adaptation to raw corrupted features.

    \item We complement these utility comparisons with separate representation-fidelity analyses, a clean-only continuation control, adaptation-data scaling, task-agnostic recovery, and statistical and
    corruption-level diagnostics.
\end{itemize}


%% file: sec/2_related_work.tex
\section{Related Work}

\paragraph{Robust multimodal understanding under visual corruption.}
Multimodal models remain sensitive to common visual corruptions. Q-Instruct
improves low-level perception through instruction-based visual-quality
supervision \citep{wu2024qinstruct}, while Robust-U1 reconstructs corrupted
visual content and jointly reasons over corrupted and recovered observations
\citep{tang2026robustu1}. These approaches improve robustness through enhanced
visual supervision or pixel-space recovery.

\paragraph{Visual representation recovery and denoising.}
Beyond pixel-space recovery, representation-space methods recover or stabilize
visual features. Denoising Vision Transformers removes positional artifacts
from ViT features using cross-view consistency \citep{yang2024dvt}, while latent
denoising recovers clean teacher features with auxiliary reconstruction and
structural objectives \citep{parikh2026latent}. Such methods primarily evaluate
recovery through feature reconstruction, alignment, and structural consistency,
making representation fidelity a central criterion.

\paragraph{Vision--language interfaces and representation alignment.}
Modern multimodal architectures expose visual representations through different
interfaces, including continuous visual prefixes, gated cross-attention,
Q-Formers, and learned projectors \citep{tsimpoukelli2021frozen,
alayrac2022flamingo,li2023blip2,liu2023visual}. JoyAI-Image further couples a
multimodal language model and a diffusion transformer through a shared interface
for understanding and generation \citep{song2026joyai}. DeCo traces semantic
flow through different projector designs and shows that the interface affects
which visual semantics reach the language model \citep{yao2024deco}. These
studies establish that interface design shapes the usability of visual
representations.

\paragraph{Representation fidelity versus downstream utility.}
Prior work establishes separate roles for feature recovery and interface design,
but it remains unclear whether a higher-fidelity recovered representation is
more useful to a fixed clean-trained readout. We study this question by
separating representation recovery from interface adaptation.


%% file: sec/3_methodology.tex
\section{Methodology}
\label{sec:method}

\begin{figure*}[t]
    \centering
    \includegraphics[width=\textwidth]{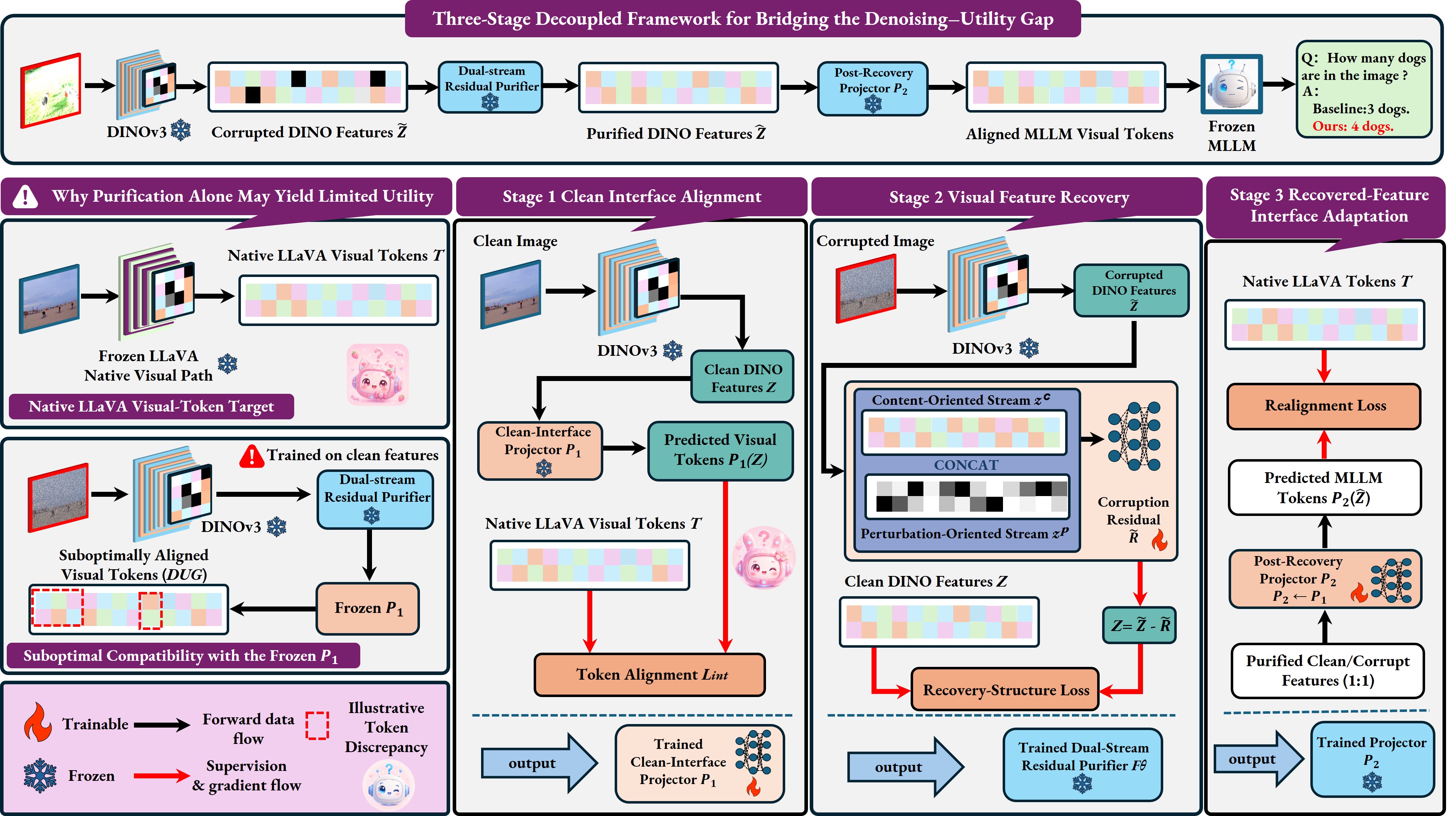}
    \caption{Overview of the controlled framework. The left panel motivates why
    improved feature fidelity may not improve downstream utility under a fixed
    interface. The three stages align a clean interface, recover corrupted
    DINOv3 features, and adapt the downstream interface to recovered features.}
    \label{fig:framework}
\end{figure*}

We study how representation recovery and readout adaptation affect
multimodal prediction. Our controlled system consists of a shared
visual encoder, a model-specific readout, and a frozen language-side
model. We intervene on the representation supplied to the readout
and on the readout parameters, while measuring the resulting
feature fidelity and downstream utility.
Figure~\ref{fig:framework} summarizes this design.

\subsection{Problem Formulation and DUG Definition}
\label{sec:problem_formulation}

We consider multimodal robustness under visual distribution shifts.
Given a clean image $x$, a corruption process $\mathcal{C}_{c,s}(\cdot)$
generates a corrupted observation:

\begin{equation}
\tilde{x}=\mathcal{C}_{c,s}(x),
\end{equation}

where $c$ denotes the corruption type and $s$ denotes the corruption severity.

A frozen visual encoder $E$ maps clean and corrupted observations into a shared
representation space:

\begin{equation}
z=E(x), \qquad
\tilde z=E(\tilde x),
\end{equation}

where $z,\tilde z\in\mathbb{R}^{N\times d}$.

In our controlled experiments, we use DINOv3 as the shared visual encoder.
Keeping the encoder fixed allows us to isolate the influence of representation
changes from variations caused by different visual backbones.

A representation recovery module $R_{\theta}$ aims to restore corrupted
features toward their clean counterparts:

\begin{equation}
\hat z=R_{\theta}(\tilde z).
\end{equation}

We distinguish representation fidelity from downstream utility.

Representation fidelity measures how close the recovered representation is to
the clean reference:

\begin{equation}
\mathcal{F}(R_{\theta})
=
\phi(\hat z,z),
\end{equation}

We use a higher-is-better convention for $\mathcal F$: for example,
$\phi$ can be cosine similarity or negative reconstruction error.
Tables report MSE in its usual lower-is-better form and mark metric
directions explicitly.

However, high representation fidelity does not necessarily imply high
multimodal utility. For an MLLM $m$, let $P^{(m)}$ denote the vision--language
interface and $L^{(m)}$ denote the frozen language-side computation. The
downstream utility is defined as:

\begin{equation}
\mathcal{U}^{(m)}(R,P)
=
S
\left(
L^{(m)}
(P(R(\tilde z))),
a
\right),
\end{equation}

where $S(\cdot)$ is the benchmark-specific evaluation metric.

The question is whether an increase in representation fidelity is
accompanied by an increase in utility under a fixed readout:

\begin{equation}
\Delta\mathcal{F}>0
\Rightarrow
\Delta\mathcal{U}>0 .
\end{equation}

We use \emph{Denoising--Utility Gap (DUG)} to denote an improvement
in representation fidelity without a corresponding improvement
in downstream utility under a fixed readout. Establishing this
pattern requires fidelity and utility measurements for the same
recovery and evaluation setting. DUG describes a discrepancy
between these outcomes, rather than a numerical difference
between their metrics.

\subsection{Controlled Representation Recovery}
\label{sec:representation_recovery}

To study the relationship between representation recovery and downstream
utility, we train recovery modules using paired clean and corrupted visual
representations without downstream multimodal supervision.

The recovery module maps a corrupted representation to a recovered
representation,

\begin{equation}
\hat z = R_{\theta}(\tilde z).
\end{equation}

A common restoration component of the recovery objective is

\begin{equation}
\mathcal{L}_{\mathrm{rest}}
=
\lambda_1 \|\hat z-z\|_2^2
+
\lambda_2 \bigl(1-\cos(\hat z,z)\bigr).
\end{equation}

Different recovery configurations may augment this restoration component with
additional structure-preserving regularization. In particular, the
task-agnostic Open Images recovery configuration uses the auxiliary terms
reported in Appendix~\ref{tab:generic_recovery_loss}.

Recovery modules are trained with representation-level supervision
and no downstream question--answer labels. For each recovery setting,
the trained module is held fixed during readout adaptation and
downstream evaluation. Representation fidelity is evaluated
separately for each setting.

\subsection{Readout Adaptation}
\label{sec:interface_diagnosis}

We test readout dependence by adapting the readout while holding
the visual encoder and recovery module fixed. This intervention
measures how updating the readout changes the utility of a given
representation.

For each evaluated MLLM $m$, we construct a model-specific replacement readout $P^{(m)}$ that maps the shared DINOv3 representation
space into the native visual-token space expected by the frozen downstream
model:

\begin{equation}
P^{(m)}:\mathcal{Z}\rightarrow\mathcal{T}^{(m)} .
\end{equation}

We first train a clean-trained interface using clean visual representations:

\begin{equation}
P_f^{(m)}
=
\arg\min_P
\ell
(P(z),t^{(m)}),
\end{equation}

Here, $t^{(m)}$ denotes the native visual-token target supplied by
the original MLLM visual pathway. In the main GQA experiments, this
target is extracted from the corresponding clean image.

The clean-trained readout serves as the fixed reference condition.
Each adapted readout is initialized from the corresponding
clean-trained readout and optimized on its adaptation distribution:
\begin{equation}
P_a^{(m)}
=
\arg\min_P
\mathbb{E}_{(z',t^{(m)})\sim\mathcal{D}_{\mathrm{adapt}}}
\left[
\ell\bigl(P(z'),t^{(m)}\bigr)
\right].
\end{equation}

In the GQA experiments, the adaptation inputs for each image pair are
\[
\begin{aligned}
\text{Raw-Adapted:}&\\
& z' \in \{E(x), E(\tilde x)\},\\
\text{Recovered-Adapted:}&\\
& z' \in \{R_{\theta}(E(x)), R_{\theta}(E(\tilde x))\},\\
\text{Clean-only Adapted:}&\\
& z' = R_{\theta}(E(x)).
\end{aligned}
\]

All three continuation conditions use clean-image native visual
tokens as targets and start from the same clean-trained readout.
The clean-only condition changes the adaptation input distribution
by excluding corrupted inputs.

During interface adaptation, all upstream visual encoders, recovery modules,
and language models remain frozen. Thus, adaptation only changes how the
downstream model consumes a given representation distribution.

\subsection{Controlled Representation--Readout Analysis}
\label{sec:controlled_diagnosis}

We evaluate a $2\times2$ design that varies the representation
supplied to the readout and whether the readout is adapted.
The representation is either the encoder output directly
(\emph{Raw}) or the output of the recovery module
(\emph{Recovered}); the readout is either held at its clean-trained
parameters (\emph{Frozen}) or adapted to the corresponding input
distribution (\emph{Adapted}). This yields Raw-Frozen,
Recovered-Frozen, Raw-Adapted, and Recovered-Adapted.
The following contrasts distinguish the utility effects of
recovery, readout adaptation, and their combination.

\paragraph{Recovery effect.}

Comparing Recovered-Frozen and Raw-Frozen measures the downstream utility
effect of inserting the recovery intervention while keeping the readout fixed:

\begin{equation}
\begin{aligned}
\Delta_{\mathrm{rec}}
&=\mathcal{U}(\text{Recovered-Frozen})\\
&\quad-\mathcal{U}(\text{Raw-Frozen}).
\end{aligned}
\end{equation}

A non-positive point estimate of $\Delta_{\mathrm{rec}}$ indicates
no observed utility gain under the fixed readout. Interpreting
this result as DUG additionally requires improved fidelity in
the same recovery and evaluation setting. Statistical uncertainty
is assessed separately in Appendix~\ref{app:statistics}.

\paragraph{Interface adaptation effect.}

Comparing Recovered-Adapted and Recovered-Frozen holds the recovered
representation fixed while changing only the downstream readout:

\begin{equation}
\begin{aligned}
\Delta_{\mathrm{adapt}}
&=\mathcal{U}(\text{Recovered-Adapted})\\
&\quad-\mathcal{U}(\text{Recovered-Frozen}).
\end{aligned}
\end{equation}

Because the recovered representation is unchanged, this comparison measures
the utility effect of changing the readout. It does not, by itself,
distinguish distribution matching from benefits due to additional readout
optimization; we evaluate this confound with a matched continuation control in
Section~\ref{sec:dug_analysis}.

\paragraph{Marginal recovery effect.}

Finally, comparing Recovered-Adapted and Raw-Adapted evaluates whether explicit
representation recovery provides additional benefit once both representations
are paired with adapted interfaces:

\begin{equation}
\begin{aligned}
\Delta_{\mathrm{marg}}
&=\mathcal{U}(\text{Recovered-Adapted})\\
&\quad-\mathcal{U}(\text{Raw-Adapted}).
\end{aligned}
\end{equation}

These contrasts distinguish changes to the representation from
changes to the readout. They quantify readout-dependent utility
without uniquely attributing the observed effects to information
loss or distribution mismatch.



%% file: sec/4_experiments.tex
\section{Experiments}
\label{sec:experiments}

Our experiments examine three questions: how recovery affects
utility under a fixed readout; how adapting the readout changes
the utility of raw and recovered features; and whether explicit
recovery remains beneficial after adaptation. We evaluate these
comparisons across four model architectures and nine benchmarks,
then examine a clean-only continuation control, adaptation-data
scale, and the recovery-training distribution. Separate
representation-level analyses characterize fidelity.

\subsection{Experimental Setup}
\label{sec:experimental_setup}

We evaluate representative multimodal large language models, including
LLaVA-1.5, Qwen2.5-VL, Qwen3.5, and LLaVA-OneVision-2, which cover
different vision--language coupling mechanisms.

A DINOv3 visual encoder defines the shared visual representation space.
The recovery module is trained to map corrupted representations toward their
clean counterparts without downstream task supervision. During evaluation,
the recovery module remains frozen and is applied before the vision--language
interface.

We consider common visual degradations including noise, blur, compression,
and exposure variation. Downstream utility is evaluated on GQA, TextVQA,
TallyQA, POPE, VQAv2, VizWiz \citep{gurari2018vizwiz}, MMStar
\citep{chen2024mmstar}, BLINK \citep{fu2024blink}, and NaturalBench
\citep{li2024naturalbench}.

We use the four representation--readout conditions defined in
Section~\ref{sec:controlled_diagnosis} and summarized in
Table~\ref{tab:condition}.

For controlled cross-model evaluation, each architecture is connected to the
shared DINOv3 representation through a separately trained diagnostic
replacement readout. The readout maps DINOv3 features into the native
visual-token space expected by the frozen downstream multimodal model; the
native visual backbone, native multimodal components, and language-side model
remain frozen. Consequently, cross-model results are interpreted through
within-model condition differences rather than absolute native benchmark
performance across MLLMs.

\subsection{Main Results and Generalization}
\label{sec:main_results}

We first compare the utility effects of recovery and readout
adaptation across models and benchmarks.

\begin{figure*}[t]
\centering
\includegraphics[width=\linewidth]{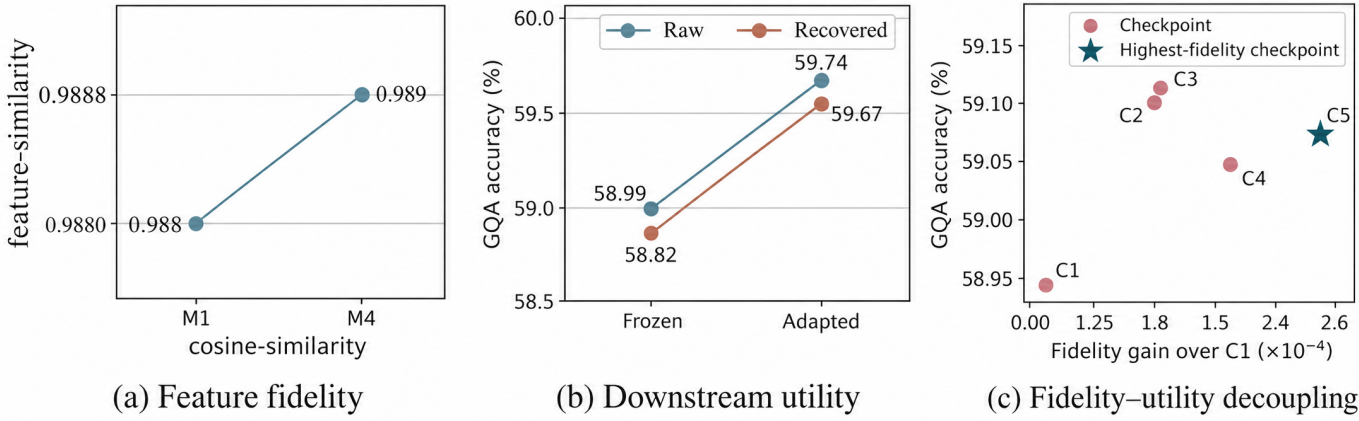}
\caption{
Representation recovery and readout adaptation affect downstream
utility through different interventions. The four conditions
compare raw and recovered features under frozen and adapted
readouts. Recovery has heterogeneous utility effects under a
fixed readout, while adaptation can change the utility of the
same representation.
}
\label{fig:dug_teaser}
\end{figure*}

Table~\ref{tab:main_results} reports the four controlled conditions across
four models and nine benchmarks. Recovery under a frozen readout does not
yield a consistent utility gain; its effect varies across tasks and
architectures.

Changing the readout produces a substantially more consistent effect.
Holding the recovered input fixed, Recovered-Adapted outperforms
Recovered-Frozen in 33 of 36 model--benchmark combinations. Because the
representation is unchanged within each comparison, this result establishes
substantial readout dependence of downstream utility.

Explicit recovery, however, provides no uniform advantage once the readout is
adapted. Recovered-Adapted exceeds Raw-Adapted in only 10 of 36 combinations;
direct adaptation to raw corrupted features is competitive with or better
than recovery followed by adaptation in the remaining cases. These results
motivate evaluating the representation--readout pair rather than judging a
recovery intervention from representation fidelity alone. The counts summarize
point estimates, not statistically significant effects.

\begin{table*}[!htbp]
\centering

\caption{
Downstream utility across four models and nine benchmarks. Scores follow
Appendix~\ref{app:benchmark_protocol} and characterize controlled systems
using a shared DINOv3 encoder and model-specific replacement readouts.
The best and second-best results within each setting are highlighted.
}
\label{tab:main_results}

\scriptsize
\setlength{\tabcolsep}{3pt}

\resizebox{\textwidth}{!}{%
\begin{tabular}{llccccccccc}

\toprule

Model &
Condition &
GQA &
TextVQA &
TallyQA &
POPE &
VQAv2 &
VizWiz &
MMStar &
BLINK &
NaturalBench
\\

\midrule


\multirow{4}{*}{LLaVA-1.5}

& Raw-Frozen
& 58.99
& 9.96
& 54.00
& \second{82.42}
& 65.45
& 17.83
& 27.53
& \second{36.12}
& 46.86
\\

& Recovered-Frozen
& 58.82
& 9.57
& 49.01
& 80.91
& 64.62
& 17.29
& 27.38
& \best{36.14}
& 45.69
\\

& Raw-Adapted
& \best{59.74}
& \best{10.11}
& \best{55.29}
& \best{82.47}
& \best{66.71}
& \best{18.38}
& \best{28.16}
& 35.77
& \best{47.42}
\\

& Recovered-Adapted
& \second{59.67}
& \second{9.97}
& \second{54.74}
& 82.18
& \second{66.53}
& \second{18.28}
& \second{27.71}
& 35.12
& \second{46.92}
\\

\midrule


\multirow{4}{*}{Qwen2.5-VL}

& Raw-Frozen
& 48.34
& 7.62
& 53.97
& 75.24
& 60.83
& 17.23
& 34.22
& 38.52
& 48.45
\\

& Recovered-Frozen
& 47.48
& 6.82
& 48.62
& 74.73
& 57.50
& 13.72
& 33.84
& 37.82
& 46.59
\\

& Raw-Adapted
& \second{49.66}
& \best{8.70}
& \second{55.38}
& \second{75.87}
& \best{63.40}
& \best{20.06}
& \best{35.53}
& \second{38.79}
& \best{49.61}
\\

& Recovered-Adapted
& \best{49.84}
& \second{8.17}
& \best{55.82}
& \best{76.17}
& \second{62.97}
& \second{18.43}
& \second{35.24}
& \best{39.05}
& \second{49.28}
\\

\midrule


\multirow{4}{*}{Qwen3.5}

& Raw-Frozen
& 54.10
& 10.87
& 60.29
& \second{83.18}
& 68.63
& 20.24
& 39.49
& \second{41.47}
& 62.79
\\

& Recovered-Frozen
& 53.79
& 9.98
& 58.29
& 82.87
& 68.05
& 17.88
& 36.91
& 41.14
& 61.47
\\

& Raw-Adapted
& \second{54.72}
& \second{11.17}
& \best{60.84}
& 83.14
& \best{69.59}
& \best{21.46}
& \best{40.44}
& \best{42.22}
& \best{63.30}
\\

& Recovered-Adapted
& \best{54.78}
& \best{11.23}
& \second{60.76}
& \best{83.42}
& \second{69.37}
& \second{20.75}
& \second{39.64}
& 41.07
& \second{63.18}
\\

\midrule


\multirow{4}{*}{OneVision-2}

& Raw-Frozen
& 19.76
& 10.11
& 32.77
& 77.16
& 23.36
& 11.30
& 34.44
& 24.88
& 46.77
\\

& Recovered-Frozen
& \second{21.05}
& 10.19
& 33.99
& 77.01
& 24.57
& 10.74
& 33.42
& \second{25.58}
& 44.50
\\

& Raw-Adapted
& 20.93
& \best{10.44}
& \second{35.64}
& \best{77.84}
& \second{26.05}
& \best{12.32}
& \best{35.80}
& \best{26.55}
& \best{47.71}
\\

& Recovered-Adapted
& \best{21.09}
& \second{10.38}
& \best{36.67}
& \second{77.73}
& \best{26.19}
& \second{11.54}
& \second{35.78}
& 25.43
& \second{47.16}
\\

\bottomrule

\end{tabular}%
}

\end{table*}

\begin{figure*}[t]
    \centering

    \begin{subfigure}[t]{0.49\textwidth}
        \centering
        \includegraphics[width=\linewidth]{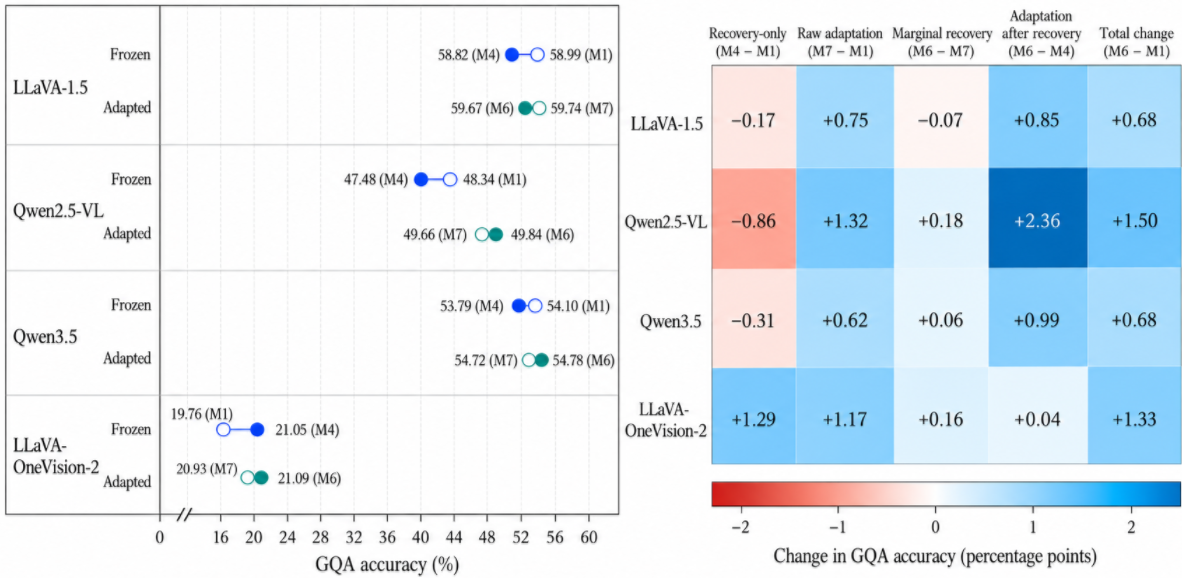}
        \caption{Cross-model utility effects of recovery and readout adaptation.}
        \label{fig:cross_model}
    \end{subfigure}
    \hfill
    \begin{subfigure}[t]{0.49\textwidth}
        \centering
        \includegraphics[width=0.49\linewidth]{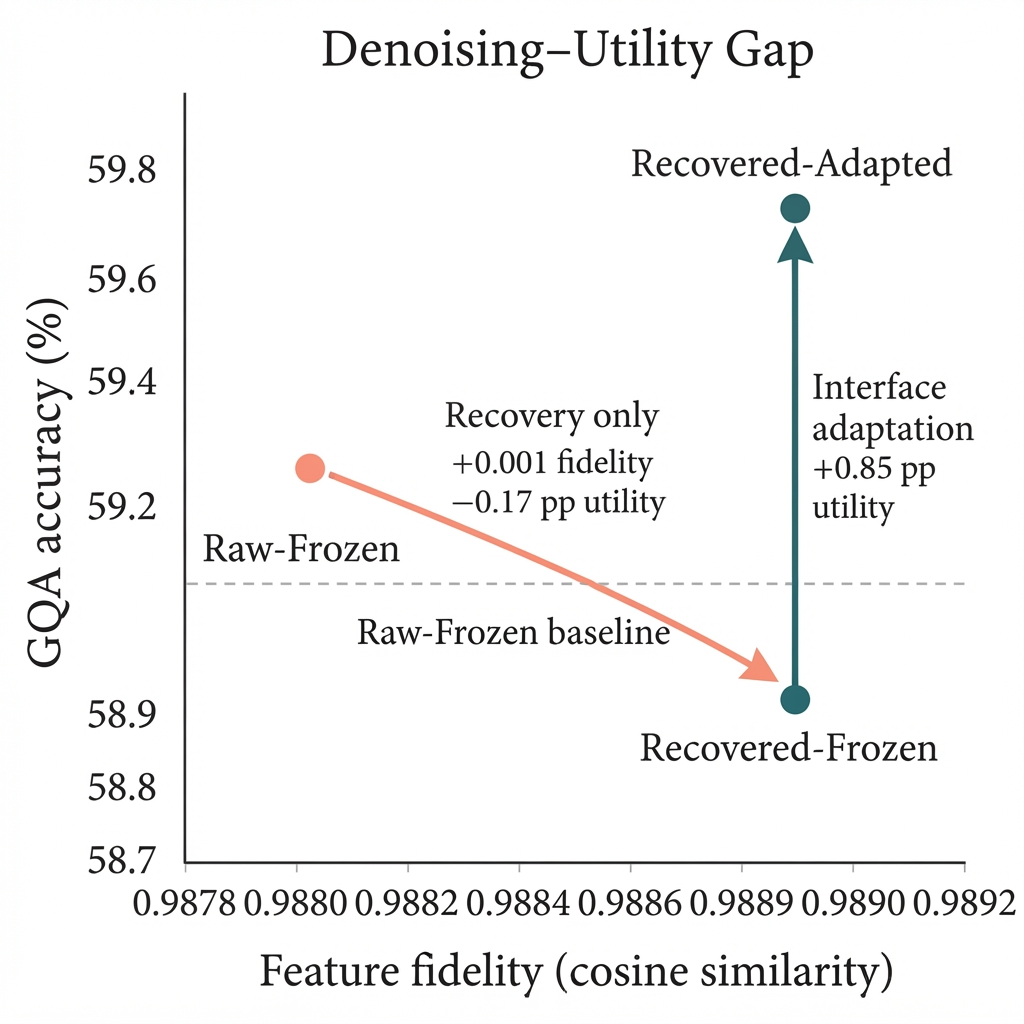}
        \caption{Controlled representation--readout comparisons.}
        \label{fig:diagnosis}
    \end{subfigure}

    \caption{
        Controlled representation--readout comparisons.
        (a) Recovery-only and recovered-input adaptation effects across
        downstream models.
        (b) Utility changes between the evaluated representation--readout
        conditions. These utility comparisons do not by themselves establish
        an improvement in representation fidelity.
    }
    \label{fig:dug_analysis}
\end{figure*}

\paragraph{Across multimodal architectures.}
We further examine whether the observed behavior is specific to a single MLLM.
Figure~\ref{fig:cross_model} decomposes the recovery-only and
interface-adaptation effects across four architectures. On GQA, the
recovery-only effect is negative for LLaVA-1.5, Qwen2.5-VL, and Qwen3.5,
but positive for OneVision-2, indicating that the utility effect of
representation recovery varies across the evaluated architectures. In contrast, adapting the readout to the recovered representation produces
a positive GQA change for all four evaluated architectures, although the
magnitude differs substantially across models.

Together, the task-level and cross-model results show that the
utility effect of recovery varies across tasks and architectures
and depends on the readout paired with the representation.

\subsection{Readout Dependence and Matched-Optimization Control}
\label{sec:dug_analysis}

Fidelity varies across recovery settings. The task-specific module
does not improve encoder-space fidelity on the GQA diagnostic
subset (Appendix~\ref{app:interface}), whereas the Open Images-trained
module improves all four fidelity metrics on ImageNet-1K
(Appendix~\ref{app:imagenet_representation}). These measurements
characterize different recovery and evaluation settings.

To distinguish distribution matching from additional readout optimization,
Table~\ref{tab:matched_optimization} adds a clean-only continuation control to
the canonical LLaVA-1.5 GQA setting.

\begin{table*}[t]
\centering
\caption{
Matched-optimization control on GQA. Clean-only adaptation excludes recovered
corrupted inputs; scores are percentages.
}
\label{tab:matched_optimization}
\small
\renewcommand{\arraystretch}{1.25}
\setlength{\tabcolsep}{4pt}
\begin{tabularx}{\linewidth}{
@{}
>{\raggedright\arraybackslash}p{0.22\linewidth}
>{\raggedright\arraybackslash}X
>{\centering\arraybackslash}p{0.09\linewidth}
>{\raggedleft\arraybackslash}p{0.20\linewidth}
@{}
}
\toprule
\textbf{Condition} &
\textbf{Adaptation input} &
\textbf{GQA $\uparrow$} &
\textbf{$\Delta$ vs. Frozen} \\
\midrule
Recovered-Frozen & Frozen & 58.82 & 0.00 \\
Clean-only Adapted & \second{Recovered clean} & \second{59.35} & \second{+0.53} \\
Recovered-Adapted & \cellcolor{bestblue}Recovered clean + corrupted & \best{59.67} & \best{+0.85} \\
\bottomrule
\end{tabularx}
\end{table*}

With the same initialization and optimization budget, clean-only adaptation
raises GQA from $58.82$ to $59.35$; including recovered corrupted inputs
reaches $59.67$. Thus, the full $0.85$-point gain cannot be attributed solely
to matching the corrupted-input distribution.


\begin{table*}[!htbp]
\centering

\caption{
Effect of interface-adaptation data scale on downstream utility.
The $0\%$ condition uses the original frozen interface, while nonzero ratios
adapt the interface using increasing fractions of distribution-matched data.
All values are percentages. Within each representation row, the best result
is highlighted in darker blue and bolded, while the second-best result is
highlighted in lighter blue.
}
\label{tab:adaptation_scaling}

\scriptsize

\begin{minipage}[t]{0.49\textwidth}
\centering

{\scriptsize\textbf{(a) GQA}}

\vspace{1mm}

\setlength{\tabcolsep}{2.6pt}

\begin{tabular}{lcccccc}

\toprule

&
\multicolumn{6}{c}{\textbf{Adaptation Data}}
\\

\cmidrule(lr){2-7}

Representation
&
$0\%$
&
$1\%$
&
$5\%$
&
$10\%$
&
$25\%$
&
$100\%$
\\

\midrule

Raw
&
58.99
&
55.85
&
59.32
&
\best{59.65}
&
\second{59.55}
&
59.44
\\

Recovered
&
58.82
&
56.87
&
59.04
&
\second{59.24}
&
59.23
&
\best{59.65}
\\

\bottomrule

\end{tabular}

\end{minipage}
\hspace{0.01\textwidth}
\begin{minipage}[t]{0.49\textwidth}
\centering

{\scriptsize\textbf{(b) POPE}}

\vspace{1mm}

\setlength{\tabcolsep}{2.6pt}

\begin{tabular}{lcccccc}

\toprule

&
\multicolumn{6}{c}{\textbf{Adaptation Data}}
\\

\cmidrule(lr){2-7}

Representation
&
$0\%$
&
$1\%$
&
$5\%$
&
$10\%$
&
$25\%$
&
$100\%$
\\

\midrule

Raw
&
\second{82.42}
&
77.17
&
81.88
&
82.10
&
82.16
&
\best{82.52}
\\

Recovered
&
80.91
&
78.47
&
81.94
&
81.90
&
\second{82.15}
&
\best{82.51}
\\

\bottomrule

\end{tabular}

\end{minipage}

\end{table*}

\begin{table*}[!htbp]
\centering

\caption{
Generalization across recovery training distributions.
The recovery module is trained either on task-specific data or on a
task-agnostic Open Images subset. We report downstream utility for recovered
representations before and after adapting the vision--language interface.
All values are percentages. Within each recovery-training setting, the better
of the two interface conditions is highlighted in light blue and bolded.
}
\label{tab:generic_openimages}

\scriptsize

\begin{tabular*}{\textwidth}{@{\extracolsep{\fill}}llccccccc}

\toprule

Recovery Training &
Condition &
GQA &
POPE &
VQAv2 &
VizWiz &
MMStar &
BLINK &
NaturalBench
\\

\midrule


\multirow{2}{*}{Task-specific}

& Recovered-Frozen
& 58.82
& 80.91
& 64.62
& 17.29
& 27.38
& \cellcolor{secondblue}\textbf{36.14}
& 45.69
\\

& Recovered-Adapted
& \cellcolor{secondblue}\textbf{59.67}
& \cellcolor{secondblue}\textbf{82.18}
& \cellcolor{secondblue}\textbf{66.53}
& \cellcolor{secondblue}\textbf{18.28}
& \cellcolor{secondblue}\textbf{27.71}
& 35.12
& \cellcolor{secondblue}\textbf{46.92}
\\

\midrule


\multirow{2}{*}{OpenImages Generic}

& Recovered-Frozen
& 59.09
& 82.37
& 65.29
& 17.79
& 27.24
& 35.82
& 47.03
\\

& Recovered-Adapted
& \cellcolor{secondblue}\textbf{59.84}
& \cellcolor{secondblue}\textbf{82.51}
& \cellcolor{secondblue}\textbf{66.63}
& \cellcolor{secondblue}\textbf{18.30}
& \cellcolor{secondblue}\textbf{28.78}
& \cellcolor{secondblue}\textbf{36.17}
& \cellcolor{secondblue}\textbf{47.36}
\\

\bottomrule

\end{tabular*}

\end{table*}

\subsection{Robustness Across Recovery Settings}
\label{sec:robustness}

We next examine the scope of the readout-adaptation results
along two dimensions: the amount of adaptation data and the
distribution used to train the recovery module. These comparisons
test whether the observed utility effects persist with smaller
adaptation sets and with task-agnostic recovery.


\paragraph{Interface adaptation with limited data.}
We vary the fraction of adaptation data while fixing the visual
representation, downstream model, and optimization budget.
The $0\%$ condition uses the frozen readout; nonzero fractions
use progressively larger subsets of the same adaptation
distribution. The $100\%$ scores in this scaling study differ
from the adapted scores in Table~\ref{tab:main_results}, so
nonzero fractions are compared within the scaling study.

On both GQA and POPE, the $1\%$ setting underperforms the
corresponding frozen baselines. Larger subsets improve utility
relative to this setting, although performance is not monotonic
in data scale. For example, raw-input GQA reaches $59.65$ at
$10\%$, compared with $59.44$ at $100\%$.

The amount of data needed to exceed the frozen baseline also
depends on the representation and task. At $5\%$ on POPE, the
raw-input score of $81.88$ remains below Raw-Frozen ($82.42$),
while the recovered-input score of $81.94$ exceeds
Recovered-Frozen ($80.91$). Thus, smaller adaptation sets can
improve utility, but do not uniformly match or exceed the
frozen baselines.


\paragraph{Task-specific versus task-agnostic recovery.}
We evaluate a recovery module trained on an Open Images subset
\citep{kuznetsova2020openimages} without downstream
question--answer supervision. On held-out ImageNet-1K
representations, it improves all four fidelity metrics
(Appendix~\ref{app:imagenet_representation}).

Under the frozen readout, downstream utility changes are mixed:
GQA and NaturalBench improve slightly, while the other five
benchmarks decrease relative to Raw-Frozen. Readout adaptation
increases all seven benchmark scores
(Table~\ref{tab:generic_openimages}), including GQA
($59.09\rightarrow59.84$), VQAv2 ($65.29\rightarrow66.63$),
and MMStar ($27.24\rightarrow28.78$). The utility gains from
readout adaptation therefore extend to recovery trained on
task-agnostic data.

The ImageNet fidelity evaluation and downstream utility
evaluations use different datasets. They characterize the same
recovery module but do not constitute matched fidelity--utility
measurements.

The purifier-design, corruption-specific, multi-seed, bootstrap,
recovery--utility sweep, and additional interface-compatibility analyses are
reported in the appendix.


%% file: sec/5_conclusion.tex
\section{Conclusion}

We study representation recovery and readout adaptation as separate
interventions in shared-encoder multimodal systems. Recovery under a
clean-trained readout has heterogeneous utility effects, whereas readout
adaptation improves utility in most evaluated settings.
Direct adaptation to raw features is frequently competitive with or
better than adaptation after explicit recovery. The clean-only continuation control shows that part of the adaptation gain is also obtained without corrupted training inputs. Separate fidelity analyses characterize
the task-specific and task-agnostic recovery settings. These results
motivate evaluating representations together with their downstream
readouts and reporting fidelity, fixed-readout utility, and
adapted-readout utility separately.


%% file: sec/appendix.tex
\section{Experimental and Reproducibility Details}
\label{app:experimental_details}

We provide additional details of the controlled representation--interface
evaluation protocol, model-specific readouts, optimization settings, benchmark
evaluation procedures, and data construction used throughout the paper.
Unless otherwise specified, downstream results use seed 123. Additional
multi-seed and statistical analyses are reported in
Appendix~\ref{app:statistics}.

\subsection{Controlled Representation--Interface Conditions}
\label{app:conditions}

Our experiments independently vary two factors: whether the visual
representation is raw or recovered, and whether the vision--language
interface remains frozen or is adapted to the corresponding representation
distribution. This produces the four semantic conditions summarized in
Table~\ref{tab:condition}.

\begin{table}[!htbp]
\centering

\caption{
Controlled representation--interface evaluation design.
Representation recovery and interface adaptation are varied independently.
}
\label{tab:condition}

\small

\begin{tabular*}{\textwidth}{@{\extracolsep{\fill}}lcc@{}}
\toprule

Representation &
Frozen Interface &
Adapted Interface
\\

\midrule

Raw &
Raw-Frozen &
Raw-Adapted
\\

Recovered &
Recovered-Frozen &
Recovered-Adapted
\\

\bottomrule
\end{tabular*}

\end{table}

\subsection{Shared Representation and Model-Specific Readouts}
\label{app:model_specific_readouts}

All evaluated architectures consume a shared DINOv3 representation containing
576 patch tokens with feature dimension 1024. Special and register tokens are
removed before the representation is passed to the model-specific
vision--language readout. Because the native visual token spaces differ across
architectures, each MLLM uses a separately trained readout that maps the shared
DINOv3 representation into the visual-token space expected by its frozen
language-side computation.

For LLaVA-1.5, the target consists of 576 native visual tokens with dimension
4096, so the DINOv3 token sequence already matches the native token count.
For Qwen2.5-VL, Qwen3.5, and OneVision-2, the native number of visual tokens is
image dependent. We therefore deterministically resample the DINOv3 sequence to
the native token count before applying the model-specific readout.

\begin{table}[!htbp]
\centering

\caption{
Model-specific vision--language readouts used in the cross-model analysis.
All models consume the same $576\times1024$ DINOv3 representation.
Dynamic token counts denote image-dependent native visual-token sequences.
}
\label{tab:model_specific_readouts}

\scriptsize

\resizebox{\textwidth}{!}{%
\begin{tabular}{lcccccl}
\toprule

Model &
Native Target Stage &
Target Tokens &
Target Dim. &
Readout &
Trainable Params. &
Token Alignment
\\

\midrule

LLaVA-1.5 &
Post-native projector &
576 &
4096 &
$1024\!\rightarrow\!4096\!\rightarrow\!4096$ &
20.98M &
Direct 576-token alignment
\\

Qwen2.5-VL &
Native visual output &
Dynamic &
3584 &
$1024\!\rightarrow\!4096\!\rightarrow\!3584$ &
18.88M &
Deterministic linear resampling
\\

Qwen3.5 &
Native visual output &
Dynamic &
4096 &
$1024\!\rightarrow\!4096\!\rightarrow\!4096$ &
20.98M &
Deterministic linear resampling
\\

OneVision-2 &
Native visual output &
Dynamic &
4096 &
$1024\!\rightarrow\!4096\!\rightarrow\!4096$ &
20.98M &
Deterministic linear resampling
\\

\bottomrule
\end{tabular}%
}

\end{table}

Each readout is a two-layer MLP with GELU activation. During interface
training and adaptation, only the readout parameters are optimized; the
DINOv3 encoder, recovery module, native multimodal model, and language model
remain frozen. Raw-Adapted and Recovered-Adapted use the same optimization
protocol within each architecture. Consequently, the cross-model experiments
are interpreted through controlled within-model differences rather than
absolute scores across architectures.

\subsection{Interface Training and Optimization}
\label{app:interface_training}

The clean-trained interface is optimized to reproduce the native visual tokens
of each MLLM from clean DINOv3 representations. The training objective combines
mean-squared error with token-wise cosine distance over valid visual tokens.
The same optimizer and optimization budget are used for all four evaluated
architectures.

\begin{table}[!htbp]
\centering

\caption{
Common optimization settings for clean-trained and adapted
vision--language readouts.
}
\label{tab:interface_training_details}

\small

\begin{tabular*}{\textwidth}{@{\extracolsep{\fill}}lc@{}}
\toprule

Setting &
Value
\\

\midrule

Optimizer & AdamW \\
Learning rate & $1\times10^{-4}$ \\
Weight decay & $0.01$ \\
Adam $\beta$ & $(0.9,\,0.95)$ \\
Gradient clipping & $1.0$ \\
Scheduler & None \\
Batch size & $64$ \\
Gradient accumulation & $1$ \\
Maximum optimization steps & $750$ \\
Minimum steps before stopping & $375$ \\
Validation subset size & $1024$ \\
Precision & bfloat16 \\
Default seed & $123$ \\
Checkpoint criterion & Minimum validation MSE + cosine loss \\

\bottomrule
\end{tabular*}

\end{table}

The clean-trained interfaces for all four architectures are trained on
72,140 clean representation--native-token pairs. Adapted readouts are
initialized from the corresponding clean-trained interface and optimized only
on the target representation distribution.

\subsection{Task-Agnostic Recovery Implementation}
\label{app:generic_recovery_impl}

For the task-agnostic Open Images experiment, we use a residual dual-stream
variational recovery module. The content branch uses a hidden dimension of
1024 and a latent dimension of 128, while the nuisance branch uses a hidden
dimension of 512 and a latent dimension of 64. The decoder hidden dimension is
1024, and the reconstructed residual is added to the input representation.
The implementation contains approximately 5.52M trainable parameters.

\begin{table}[!htbp]
\centering

\caption{
Architecture and optimization settings of the task-agnostic Open Images
recovery module.
}
\label{tab:generic_recovery_details}

\small

\begin{tabular*}{\textwidth}{@{\extracolsep{\fill}}lc@{}}
\toprule

Setting &
Value
\\

\midrule

Input / output dimension & 1024 \\
Content hidden / latent & 1024 / 128 \\
Nuisance hidden / latent & 512 / 64 \\
Decoder hidden dimension & 1024 \\
Activation & GELU \\
Layer normalization & Yes \\
Dropout & 0 \\
Approx.\ parameters & 5.52M \\
Optimizer & AdamW \\
Learning rate & $3\times10^{-4}$ \\
Weight decay & $0.01$ \\
Adam $\beta$ & $(0.9,\,0.95)$ \\
Gradient clipping & $1.0$ \\
Batch size & 32 \\
Scheduler & Cosine \\
Warmup steps & 200 \\
Training steps & 12,000 \\
Minimum early-stop step & 8,000 \\
Training seed & 123 \\

\bottomrule
\end{tabular*}

\end{table}

The task-agnostic recovery objective combines direct restoration losses with
structure-preserving regularization. Table~\ref{tab:generic_recovery_loss}
lists the loss weights used to train the Open Images recovery module.

\begin{table}[!htbp]
\centering

\caption{
Loss weights used for the task-agnostic Open Images recovery module.
}
\label{tab:generic_recovery_loss}

\small

\begin{tabular*}{\textwidth}{@{\extracolsep{\fill}}lc@{}}
\toprule

Loss Term &
Weight
\\

\midrule

Restoration MSE & 1.0 \\
Restoration cosine & 1.0 \\
Restoration Smooth-$L_1$ & 0.5 \\
Global cosine & 0.1 \\
Relational consistency & 0.2 \\
KL regularization & 0.005 \\
Total-correlation regularization & 0.02 \\
Cycle consistency & 0.3 \\
Minimum clean-identity term & 0.01 \\
Minimum residual-$L_2$ term & 0.02 \\

\bottomrule
\end{tabular*}

\end{table}

The canonical task-specific experiments use a fixed task-specific recovery
checkpoint throughout downstream evaluation. In all downstream conditions,
the recovery module is frozen; interface adaptation therefore never changes
the recovered representation itself.

\subsection{Benchmark Evaluation Protocol}
\label{app:benchmark_protocol}

We evaluate all representation--interface conditions using the same
benchmark-specific scorer within each task. Model generations are decoded only
from the continuation after the prompt. Empty generations are treated as
incorrect. Unless the benchmark requires a dedicated protocol, answers are
lowercased, punctuation and articles are removed, and whitespace is
canonicalized before scoring.

Avg-All-C is the macro-average of the available Clean, Medium, and
Heavy condition scores. Each condition score is computed over its
evaluated rows. When all three conditions are present,
\[
S_{\mathrm{Avg\mbox{-}All\mbox{-}C}}
=
\frac{
S_{\mathrm{Clean}}+
S_{\mathrm{Medium}}+
S_{\mathrm{Heavy}}
}{3}.
\]
Thus, Avg-All-C includes clean inputs and is distinct from the
corrupted-only Avg-C score. Evaluated-row counts include condition
expansion and should not be interpreted as counts of unique images.

\begin{table}[!htbp]
\centering

\caption{
Benchmark-specific scoring protocols.
TextVQA and VizWiz use the soft score
$\min(1,\mathrm{matches}/3)$, whereas VQAv2 uses normalized
exact-match accuracy. These study-specific scoring rules differ
from the corresponding official VQA metrics. Counts refer to
evaluated rows.
}
\label{tab:benchmark_protocol}

\scriptsize

\resizebox{\textwidth}{!}{%
\begin{tabular}{llrll}
\toprule

Benchmark &
Split &
Evaluated rows &
Metric &
Answer Processing
\\

\midrule

GQA &
val &
30,000 &
Normalized exact match &
Lowercase, punctuation/article removal
\\

TextVQA &
val &
15,000 &
VQA-style soft accuracy &
Standard normalized short-answer matching
\\

TallyQA &
test &
30,000 &
Exact count accuracy &
Normalization + number words $0$--$20$
\\

POPE &
val2014 &
27,000 &
Canonical yes/no accuracy &
Official first-sentence yes/no extraction
\\

VQAv2 &
screening-seed123 &
15,000 &
Implemented normalized exact match &
Normalized answer matching
\\

VizWiz &
val &
12,957 &
Implemented soft accuracy &
Standard normalized short-answer matching
\\

MMStar &
val &
4,500 &
Normalized exact match &
Lowercase, punctuation/article removal
\\

BLINK &
val &
5,703 &
Normalized exact match &
Lowercase, punctuation/article removal
\\

NaturalBench &
train &
22,800 &
Normalized exact match &
Lowercase, punctuation/article removal
\\

\bottomrule
\end{tabular}%
}

\end{table}

\paragraph{POPE answer parsing.}
POPE predictions are evaluated using the same parser across all models.
Empty generations are marked incorrect before applying the yes/no
parser to nonempty responses. For nonempty responses, only the first
sentence is considered and commas are removed. Responses containing
the tokens ``No'', ``no'', or ``not'' are classified as negative;
all other nonempty responses are classified as affirmative.
The same procedure is applied to LLaVA-1.5, Qwen2.5-VL, Qwen3.5,
and OneVision-2, with no architecture-specific parser.

\subsection{Interface-Adaptation Scaling Protocol}
\label{app:adaptation_scaling_protocol}

The adaptation-scaling experiment varies the number of unique
distribution-matched samples available to the interface while holding the
configured optimization budget fixed. The nonzero subsets are nested and are
constructed with seed 123. Raw and recovered conditions use the same example
identities at each ratio and are initialized from the same clean-trained
interface.

The $0\%$ condition reported in the main paper corresponds to the original
frozen interface and involves no adaptation training. We therefore omit it
from the adaptation-training protocol in Table~\ref{tab:adaptation_scaling_protocol}
and report only the nonzero adaptation subsets.

\begin{table}[!htbp]
\centering

\caption{
Data subsets used for interface adaptation.
All reported conditions use the same 750-step optimization budget.
}
\label{tab:adaptation_scaling_protocol}

\small

\begin{tabular*}{\textwidth}{@{\extracolsep{\fill}}rccc@{}}
\toprule

Adaptation Ratio &
Unique Samples &
Nested &
Optimization Steps
\\

\midrule

$1\%$   & 76    & Yes & 750 \\
$5\%$   & 384   & Yes & 750 \\
$10\%$  & 769   & Yes & 750 \\
$25\%$  & 1,922 & Yes & 750 \\
$100\%$ & 7,691 & Yes & 750 \\

\bottomrule
\end{tabular*}

\end{table}

All adaptation settings use AdamW with learning rate $10^{-4}$ and batch
size 64. Because the optimization budget is fixed across ratios, the experiment
varies the diversity of available adaptation examples rather than allocating
more optimization steps to larger subsets.

\subsection{Corruption Generation Protocol}
\label{app:corruption_protocol}

We evaluate five common corruption families at three severity levels. The
corruption instances are deterministic and shared across evaluated models.
Gaussian-noise randomness is seeded from a hash of the protocol identifier,
sample identifier, corruption family, and severity.

\begin{table}[!htbp]
\centering

\caption{
Parameters used to construct the three corruption severity levels.
For exposure corruptions, the value denotes the multiplicative brightness
factor.
}
\label{tab:corruption_parameters}

\small

\begin{tabular*}{\textwidth}{@{\extracolsep{\fill}}lccc@{}}
\toprule

Corruption &
Severity 1 &
Severity 2 &
Severity 3
\\

\midrule

Gaussian Blur (radius) &
1.0 & 2.0 & 4.0
\\

Gaussian Noise ($\sigma$) &
8 & 16 & 32
\\

JPEG (quality) &
70 & 45 & 25
\\

Overexposure (factor) &
1.35 & 1.75 & 2.25
\\

Underexposure (factor) &
0.75 & 0.55 & 0.35
\\

\bottomrule
\end{tabular*}

\end{table}

Training excludes the held-out heavy corruption settings used for final test
evaluation. Downstream labels are not used to train either the recovery module
or the representation--interface readout.

\subsection{Task-Agnostic Open Images Recovery Data}
\label{app:openimages_protocol}

We train the task-agnostic recovery module on paired clean and
corrupted DINOv3 features from 72,140 images in Open Images. Images are assigned using a deterministic
SHA256-sorted balanced round-robin policy. Training uses paired features without downstream
question--answer annotations.

Clean and corrupted feature pairs are generated using the same corruption
protocol as the task-specific setting, and the clean DINOv3 representation is
used as the recovery target. The recovery model is trained for 12,000 steps
with seed 123. No GQA, POPE, VQAv2, or other downstream QA annotations are
used during this training.

\begin{table}[!htbp]
\centering

\caption{
Task-agnostic Open Images recovery-data protocol.
}
\label{tab:openimages_protocol}

\small

\begin{tabular*}{\textwidth}{@{\extracolsep{\fill}}lc@{}}
\toprule

Setting &
Value
\\

\midrule

Images & 72,140 \\
Supervision & Clean/corrupted feature pairs \\
QA supervision & None \\
Training seed & 123 \\
Corruption seed & 20260920 \\
Recovery steps & 12,000 \\
Downstream annotations used & No \\
ImageNet-1K validation images & 50,000 \\

\bottomrule
\end{tabular*}

\end{table}

We evaluate this module on 50,000 ImageNet-1K validation images
\citep{deng2009imagenet} to measure representation fidelity on a
separate dataset (Appendix~\ref{app:imagenet_representation}).

\section{Recovery Design and Training Controls}
\label{app:recovery_ablation}

We examine how recovery architecture and training objective
affect representation fidelity and downstream utility, and
assess additional readout optimization using a clean-only
continuation control.

\subsection{Recovery Architecture and Objective Ablation}
\label{app:purifier_design}

Table~\ref{tab:purifier_ablation_full} compares alternative recovery designs
using both representation-level fidelity metrics and downstream utility.
The full recovery objective obtains the strongest reconstruction metrics, but
does not yield the highest downstream GQA score. This provides an additional
example that improvements in representation fidelity do not translate
monotonically into downstream utility.

\begin{table}[!htbp]
\centering

\caption{
Sensitivity to recovery architecture and training objective.
We report both representation fidelity and downstream utility.
Lower MSE and normalized MSE are better, whereas higher cosine similarity,
alignment, and downstream scores are better.
}
\label{tab:purifier_ablation_full}

\scriptsize

\begin{tabular*}{\textwidth}{@{\extracolsep{\fill}}lcccccc@{}}
\toprule

Variant &
MSE $\downarrow$ &
Norm-MSE $\downarrow$ &
Cosine $\uparrow$ &
Alignment $\uparrow$ &
GQA $\uparrow$ &
POPE $\uparrow$
\\

\midrule

Raw
& 0.001885
& 0.023075
& 0.988466
& 0.996742
& 58.99
& 82.42
\\

Linear
& 0.001975
& 0.024196
& 0.988140
& 0.996402
& 59.09
& 82.36
\\

Shallow
& 0.001944
& 0.023839
& 0.988423
& 0.996650
& \textbf{59.17}
& 82.30
\\

Rec-only
& 0.001906
& 0.023372
& 0.988560
& \textbf{0.996849}
& 59.14
& \textbf{82.41}
\\

Align-only
& 0.002035
& 0.024946
& 0.987695
& 0.995914
& 59.09
& 82.30
\\

Full
& \textbf{0.001864}
& \textbf{0.022826}
& \textbf{0.988735}
& 0.996833
& 59.09
& 82.37
\\

\bottomrule
\end{tabular*}

\end{table}

\subsection{Matched-Optimization Control}
\label{app:clean_control}

The clean-only control in Table~\ref{tab:matched_optimization}
continues readout training on recovered clean features.
Comparing this condition with Recovered-Adapted assesses the
additional utility associated with including corrupted inputs
during adaptation.

\subsection{Task-Agnostic Representation Validation}
\label{app:imagenet_representation}

We additionally evaluate the Open Images-trained recovery module on held-out
ImageNet-1K representations. This test measures whether the generic recovery
module improves representation fidelity independently of downstream multimodal
supervision.

\begin{table}[!htbp]
\centering

\caption{
Representation-level validation of the task-agnostic Open Images recovery
module on ImageNet-1K. Lower error and higher similarity are better.
}
\label{tab:imagenet_representation}

\small

\begin{tabular*}{\textwidth}{@{\extracolsep{\fill}}lcccc@{}}
\toprule

Representation &
MSE $\downarrow$ &
Norm-MSE $\downarrow$ &
Cosine $\uparrow$ &
Alignment $\uparrow$
\\

\midrule

Raw
& 0.003091
& 0.032864
& 0.982917
& 0.993639
\\

Recovered
& \textbf{0.002993}
& \textbf{0.031825}
& \textbf{0.983556}
& \textbf{0.994151}
\\

\bottomrule
\end{tabular*}

\end{table}

Recovery reduces both reconstruction-error metrics and increases
both similarity metrics on ImageNet-1K, extending its fidelity
improvements to a dataset distinct from the Open Images
training set.

\section{Statistical Reliability}
\label{app:statistics}

We evaluate the stability of the controlled comparisons across random seeds
and quantify uncertainty using paired bootstrap confidence intervals.

\subsection{Multi-Seed Evaluation}
\label{app:multiseed}

\begin{table}[!htbp]
\centering

\caption{
GQA performance across three random seeds.
Mean and standard deviation are computed over the three runs.
All values are percentages.
}
\label{tab:multiseed}

\small

\begin{tabular*}{\textwidth}{@{\extracolsep{\fill}}lcccc@{}}
\toprule

Seed &
Raw-Frozen &
Recovered-Frozen &
Raw-Adapted &
Recovered-Adapted
\\

\midrule

42
& 59.03
& 58.67
& 59.63
& 59.67
\\

123
& 58.99
& 58.82
& 59.74
& 59.67
\\

3407
& 59.18
& 58.93
& 59.45
& 59.44
\\

\midrule

Mean
& 59.07
& 58.81
& 59.61
& 59.59
\\

Std.
& 0.10
& 0.13
& 0.15
& 0.13
\\

\bottomrule
\end{tabular*}

\end{table}

Across all three seeds, Recovered-Frozen remains below Raw-Frozen, whereas
interface adaptation increases the utility of the recovered representation.
The qualitative controlled comparison is therefore stable across the evaluated
random seeds.

\subsection{Bootstrap Confidence Intervals}
\label{app:bootstrap}

We additionally estimate uncertainty for the four controlled effects on the
primary LLaVA-1.5 downstream benchmarks. The analysis uses a paired bootstrap
over the intersection of evaluation-example identifiers in the two conditions
being compared. We use 5,000 bootstrap resamples and report 95\% percentile
confidence intervals corresponding to the 2.5th and 97.5th percentiles of the
bootstrap distribution.

These intervals condition on the fitted checkpoints and describe
evaluation-sample uncertainty under the stated example-level
resampling scheme. They do not measure variability across training
seeds. Because questions and corruption instances may share an
image, example-level resampling may underrepresent within-image
dependence. The reported intervals are therefore interpreted with
this limitation.

We denote the recovery effect by $\Delta_{\mathrm{rec}}$, the
interface-adaptation effect by $\Delta_{\mathrm{adapt}}$, the raw-adaptation
effect by $\Delta_{\mathrm{raw}}$, and the marginal recovery effect after
adaptation by $\Delta_{\mathrm{marg}}$.

\begin{align}
\Delta_{\mathrm{rec}}
&=
\mathcal{U}(\text{Recovered-Frozen})
-
\mathcal{U}(\text{Raw-Frozen}),
\\
\Delta_{\mathrm{adapt}}
&=
\mathcal{U}(\text{Recovered-Adapted})
-
\mathcal{U}(\text{Recovered-Frozen}),
\\
\Delta_{\mathrm{raw}}
&=
\mathcal{U}(\text{Raw-Adapted})
-
\mathcal{U}(\text{Raw-Frozen}),
\\
\Delta_{\mathrm{marg}}
&=
\mathcal{U}(\text{Recovered-Adapted})
-
\mathcal{U}(\text{Raw-Adapted}).
\end{align}

\begin{table}[!htbp]
\centering

\caption{
Paired-bootstrap confidence intervals for the controlled utility effects on
LLaVA-1.5. Each entry reports the percentage-point difference followed by its
95\% percentile confidence interval.95\% percentile confidence interval. Differences are computed before
rounding and may differ by $0.01$ percentage points from subtracting
the displayed scores.
}
\label{tab:bootstrap_ci}

\scriptsize

\begin{tabular*}{\textwidth}{@{\extracolsep{\fill}}lcccc@{}}
\toprule

Benchmark &
$\Delta_{\mathrm{rec}}$ &
$\Delta_{\mathrm{adapt}}$ &
$\Delta_{\mathrm{raw}}$ &
$\Delta_{\mathrm{marg}}$
\\

\midrule

GQA
&
$-0.17\;[-0.44,\,0.11]$
&
$+0.85\;[0.58,\,1.14]$
&
$+0.75\;[0.53,\,0.96]$
&
$-0.07\;[-0.26,\,0.12]$
\\

TextVQA
&
$-0.39\;[-0.60,\,-0.18]$
&
$+0.40\;[0.18,\,0.62]$
&
$+0.16\;[-0.02,\,0.34]$
&
$-0.14\;[-0.29,\,0.01]$
\\

TallyQA
&
$-4.99\;[-5.41,\,-4.59]$
&
$+5.72\;[5.29,\,6.16]$
&
$+1.28\;[0.98,\,1.57]$
&
$-0.55\;[-0.79,\,-0.31]$
\\

POPE
&
$-1.51\;[-1.77,\,-1.23]$
&
$+1.26\;[1.00,\,1.52]$
&
$+0.04\;[-0.14,\,0.24]$
&
$-0.29\;[-0.46,\,-0.12]$
\\

VQAv2
&
$-0.83\;[-1.23,\,-0.43]$
&
$+1.91\;[1.47,\,2.31]$
&
$+1.26\;[0.97,\,1.57]$
&
$-0.18\;[-0.43,\,0.07]$
\\

\bottomrule
\end{tabular*}

\end{table}

For the benchmarks listed here, recovery under a frozen readout has
negative point estimates, whereas adaptation on recovered inputs
has positive point estimates. These comparisons characterize
readout-dependent utility changes without uniquely identifying
their mechanism. Small differences between adapted raw and recovered
conditions are not interpreted as evidence of equivalence.

\section{Corruption-Level Robustness}
\label{app:corruption}

We further disaggregate GQA performance by corruption family and severity.
This analysis tests whether the observed behavior is driven by a small subset
of visual degradations.

\begin{table}[!htbp]
\centering

\caption{
Corruption-specific GQA performance across corruption families and severity
levels. All values are percentages.
}
\label{tab:corruption_breakdown}

\scriptsize

\begin{tabular*}{\textwidth}{@{\extracolsep{\fill}}llcccc@{}}
\toprule

Corruption &
Severity &
Raw-Frozen &
Recovered-Frozen &
Raw-Adapted &
Recovered-Adapted
\\

\midrule

\multirow{3}{*}{Gaussian Blur}
& 1 & 59.31 & 60.20 & 60.20 & 59.80 \\
& 2 & 59.76 & 59.76 & 60.96 & 60.46 \\
& 3 & 55.28 & 54.64 & 56.12 & 56.47 \\

\midrule

\multirow{3}{*}{Gaussian Noise}
& 1 & 59.28 & 59.67 & 60.06 & 60.93 \\
& 2 & 60.86 & 60.46 & 60.66 & 61.16 \\
& 3 & 58.32 & 57.76 & 59.12 & 58.52 \\

\midrule

\multirow{3}{*}{JPEG}
& 1 & 57.57 & 56.87 & 58.57 & 58.67 \\
& 2 & 58.59 & 57.80 & 59.68 & 60.08 \\
& 3 & 58.06 & 58.45 & 58.25 & 58.55 \\

\midrule

\multirow{3}{*}{Overexposure}
& 1 & 56.83 & 57.25 & 57.35 & 57.35 \\
& 2 & 58.64 & 59.26 & 59.47 & 58.74 \\
& 3 & 60.52 & 60.62 & 62.07 & 62.37 \\

\midrule

\multirow{3}{*}{Underexposure}
& 1 & 59.59 & 59.98 & 59.59 & 59.98 \\
& 2 & 61.26 & 61.26 & 61.76 & 62.06 \\
& 3 & 60.12 & 59.52 & 60.58 & 60.27 \\

\bottomrule
\end{tabular*}

\end{table}

The effect of recovery alone varies across corruption families and severity
levels rather than producing a uniform utility gain. Adapted-interface
conditions are generally competitive with or stronger than their corresponding
frozen conditions, further supporting the distinction between representation
recovery and representation usability.

\section{Additional Fidelity--Utility Diagnostics}
\label{app:fidelity_utility}

We provide additional analyses of how representation metrics and
downstream utility vary across recovery checkpoints.

\subsection{Recovery Checkpoint Sweep}
\label{app:recovery_sweep}

Table~\ref{tab:recovery_sweep} evaluates five recovery checkpoints.
The fidelity metrics improve overall from C1 to C5,
although MSE and normalized MSE increase slightly from C2 to C3.
GQA reaches its highest score at C3 ($59.16$), while POPE
reaches its largest score at C2 ($82.40$). C5 obtains $59.09$ and
$82.37$, respectively. Thus, the checkpoint with the best final
fidelity metrics does not maximize either downstream score in
this sweep.

\begin{table}[!htbp]
\centering

\caption{
Recovery--utility sweep across five checkpoints.
Representation fidelity improves overall from C1 to C5, while
the displayed GQA and POPE scores are non-monotonic across the
checkpoint sequence. C1--C5 identify the displayed checkpoints.
}
\label{tab:recovery_sweep}

\scriptsize

\begin{tabular*}{\textwidth}{@{\extracolsep{\fill}}lcccccc@{}}
\toprule

Checkpoint &
MSE $\downarrow$ &
Norm-MSE $\downarrow$ &
Cosine $\uparrow$ &
Alignment $\uparrow$ &
GQA $\uparrow$ &
POPE $\uparrow$
\\

\midrule

C1
& 0.001894
& 0.023185
& 0.988469
& 0.996452
& 58.93
& 82.27
\\

C2
& 0.001886
& 0.023095
& 0.988594
& 0.996724
& 59.14
& 82.40
\\

C3
& 0.001887
& 0.023099
& 0.988596
& 0.996759
& \textbf{59.16}
& 82.30
\\

C4
& 0.001871
& 0.022913
& 0.988668
& 0.996777
& 59.05
& 82.34
\\

C5
& \textbf{0.001864}
& \textbf{0.022826}
& \textbf{0.988735}
& \textbf{0.996833}
& 59.09
& 82.37
\\

\bottomrule
\end{tabular*}

\end{table}

\subsection{Fidelity--Utility Correlation}
\label{app:fidelity_corr}

We summarize the association between representation metrics and downstream
utility over the same five recovery checkpoints. Because the sweep contains
only five checkpoints, these values are descriptive diagnostics rather than
formal statistical evidence.

\begin{table}[!htbp]
\centering

\caption{
Descriptive correlation between representation-recovery metrics and downstream
utility over five recovery checkpoints.
}
\label{tab:fidelity_utility_corr}

\small

\begin{tabular*}{\textwidth}{@{\extracolsep{\fill}}llcc@{}}
\toprule

Benchmark &
Recovery Metric &
Pearson &
Spearman
\\

\midrule

\multirow{4}{*}{GQA}
& MSE              & -0.25 & -0.10 \\
& Normalized MSE   & -0.24 & -0.10 \\
& Cosine           &  0.52 &  0.20 \\
& Global Alignment &  0.79 &  0.20 \\

\midrule

\multirow{4}{*}{POPE}
& MSE              & -0.52 & -0.70 \\
& Normalized MSE   & -0.51 & -0.70 \\
& Cosine           &  0.65 &  0.40 \\
& Global Alignment &  0.67 &  0.40 \\

\bottomrule
\end{tabular*}

\end{table}

The associations vary by metric and task. The reported GQA Pearson
correlation with global alignment is $0.79$, and the POPE Spearman
correlation with MSE is $-0.70$. Because lower MSE indicates higher
fidelity, the latter is directionally consistent with better fidelity
accompanying higher utility. These five-checkpoint diagnostics do
not support a general claim of no association. Their narrower
observation is that downstream utility is non-monotonic across the
displayed checkpoint sequence and is not maximized by C5.

\section{Interface Compatibility Diagnostics}
\label{app:interface}

We complement downstream evaluation with direct measurements of how each
representation distribution interacts with the corresponding vision--language
interface. We report statistics both before and after the model-specific
interface transformation.

\begin{table}[!htbp]
\centering

\caption{
Representation--interface compatibility diagnostics on the GQA interface
sample. ``Pre'' denotes DINOv3 features measured against the
corresponding clean DINOv3 reference, while ``Post'' denotes
measurements after the model-specific readout transformation.
Lower MSE and normalized MSE and higher cosine similarity are better.
}
\label{tab:interface_compatibility}

\scriptsize

\begin{tabular*}{\textwidth}{@{\extracolsep{\fill}}llccccc@{}}
\toprule

Condition &
Stage &
Cosine $\uparrow$ &
MSE $\downarrow$ &
Norm-MSE $\downarrow$ &
Pooled Cosine $\uparrow$ &
Norm Ratio
\\

\midrule

\multirow{2}{*}{Raw-Frozen}
& Pre
& 0.988435
& 0.001886
& 0.023091
& 0.996749
& 1.001307
\\

& Post
& 0.998598
& 0.001082
& 0.005069
& 0.999791
& 1.008561
\\

\midrule

\multirow{2}{*}{Recovered-Frozen}
& Pre
& 0.894237
& 0.022155
& 0.272020
& 0.826113
& 1.134325
\\

& Post
& 0.976949
& 0.011225
& 0.056457
& 0.981581
& 1.002465
\\

\midrule

\multirow{2}{*}{Raw-Adapted}
& Pre
& 0.988435
& 0.001886
& 0.023091
& 0.996749
& 1.001307
\\

& Post
& 0.992660
& 0.003378
& 0.017589
& 0.998397
& 1.021070
\\

\midrule

\multirow{2}{*}{Recovered-Adapted}
& Pre
& 0.894237
& 0.022155
& 0.272020
& 0.826113
& 1.134325
\\

& Post
& 0.992517
& 0.003464
& 0.017791
& 0.998598
& 0.997084
\\

\bottomrule
\end{tabular*}

\end{table}

On this GQA diagnostic subset, task-specific recovery lowers
token-wise cosine similarity to the clean DINOv3 reference
from $0.988435$ to $0.894237$ and increases MSE from
$0.001886$ to $0.022155$, indicating reduced encoder-space
fidelity. Readout adaptation changes the post-interface
statistics while leaving the input features unchanged.
These measurements characterize changes before and after
the readout; they do not isolate the cause of the downstream
utility changes.

\section{Additional Benchmark Analyses}
\label{app:additional_benchmarks}

\subsection{Detailed POPE Metrics}
\label{app:pope_details}

POPE accuracy alone may obscure changes in answer distribution or class
imbalance. We therefore report precision, recall, F1, and yes-answer ratio in
Table~\ref{tab:pope_full}. All conditions are evaluated with the same canonical
yes/no parser described in Appendix~\ref{app:benchmark_protocol}.

\begin{table}[!htbp]
\centering

\caption{
Detailed POPE evaluation.
All metric values except sample count are percentages.
}
\label{tab:pope_full}

\scriptsize

\begin{tabular*}{\textwidth}{@{\extracolsep{\fill}}llrrrrrr@{}}
\toprule

Source &
Condition &
$n$ &
Acc. &
Prec. &
Recall &
F1 &
Yes Ratio
\\

\midrule

Canonical
& Clean-Frozen
& 27000
& 82.42
& 87.81
& 75.30
& 81.07
& 42.87
\\

Canonical
& Raw-Frozen
& 27000
& 82.42
& 87.81
& 75.30
& 81.07
& 42.87
\\

Canonical
& Recovered-Frozen
& 27000
& 80.91
& 87.60
& 72.02
& 79.05
& 41.11
\\

Canonical
& Raw-Adapted
& 27000
& 82.47
& 88.19
& 74.98
& 81.05
& 42.51
\\

Canonical
& Recovered-Adapted
& 27000
& 82.18
& 87.88
& 74.65
& 80.73
& 42.47
\\

OpenImages
& Recovered-Frozen
& 27000
& 82.37
& 87.71
& 75.29
& 81.03
& 42.92
\\

OpenImages
& Recovered-Adapted
& 27000
& 82.51
& 87.93
& 75.36
& 81.16
& 42.85
\\

\bottomrule
\end{tabular*}

\end{table}

The yes-answer ratio ranges from $41.11\%$ to $42.92\%$ across
the evaluated conditions. For task-specific recovered inputs,
readout adaptation increases precision from $87.60\%$ to
$87.88\%$ and recall from $72.02\%$ to $74.65\%$.

\subsection{TallyQA-Complex}
\label{app:tallyqa_complex}

On TallyQA-Complex, recovery under the frozen readout lowers
accuracy from $40.17$ to $38.48$. Adapting the readout to
recovered features raises accuracy to $40.77$, exceeding
Raw-Frozen while remaining below Raw-Adapted ($41.24$).
This subset therefore also shows a benefit from readout
adaptation without an additional advantage from explicit recovery.

\begin{table}[!htbp]
\centering

\caption{
Performance on the TallyQA-Complex subset.
All values are percentages.
}
\label{tab:tallyqa_complex}

\small

\begin{tabular*}{\textwidth}{@{\extracolsep{\fill}}lc@{}}
\toprule

Condition &
TallyQA-Complex $\uparrow$
\\

\midrule

Raw-Frozen & 40.17 \\
Recovered-Frozen & 38.48 \\
Raw-Adapted & 41.24 \\
Recovered-Adapted & 40.77 \\

\bottomrule
\end{tabular*}

\end{table}

The complex counting subset exhibits the same qualitative asymmetry as the
main TallyQA evaluation: recovery alone reduces downstream utility, while
interface adaptation recovers a substantial portion of the lost performance.

\subsection{Relation to Main-Text Robustness Experiments}
\label{app:main_robustness_relation}

The main paper reports the complete task-agnostic Open Images transfer
experiment and the interface-adaptation data-scaling study. We do not duplicate
their utility tables here. The additional controls in this appendix instead
test robustness to recovery architecture, optimization, representation
fidelity, corruption family, random seed, and downstream task, while Sections
\ref{app:adaptation_scaling_protocol} and \ref{app:openimages_protocol}
provide the corresponding reproducibility details.

%% file: main.bbl
\begin{thebibliography}{20}
\providecommand{\natexlab}[1]{#1}
\providecommand{\url}[1]{\texttt{#1}}
\expandafter\ifx\csname urlstyle\endcsname\relax
  \providecommand{\doi}[1]{doi: #1}\else
  \providecommand{\doi}{doi: \begingroup \urlstyle{rm}\Url}\fi

\bibitem[Alayrac et~al.(2022)Alayrac, Donahue, Luc, Miech, et~al.]{alayrac2022flamingo}
Jean-Baptiste Alayrac, Jeff Donahue, Pauline Luc, Antoine Miech, et~al.
\newblock Flamingo: A visual language model for few-shot learning.
\newblock In \emph{Advances in Neural Information Processing Systems (NeurIPS)}, 2022.

\bibitem[Chen et~al.(2024)Chen, Li, Dong, Zhang, Zang, Chen, Duan, Wang, Yu, Lin, and Zhao]{chen2024mmstar}
Lin Chen, Jinsong Li, Xiaoyi Dong, Pan Zhang, Yuhang Zang, Zehui Chen, Haodong Duan, Jiaqi Wang, Qiao Yu, Dahua Lin, and Feng Zhao.
\newblock Are we on the right way for evaluating large vision-language models?
\newblock In \emph{Advances in Neural Information Processing Systems (NeurIPS)}, pages 27056--27087, 2024.

\bibitem[Chhipa et~al.(2023)Chhipa, Holmgren, De, Saini, and Liwicki]{chhipa2023self}
Prakash~Chandra Chhipa, Johan~Rodahl Holmgren, Kanjar De, Rajkumar Saini, and Marcus Liwicki.
\newblock Can self-supervised representation learning methods withstand distribution shifts and corruptions?
\newblock In \emph{Proceedings of the IEEE/CVF International Conference on Computer Vision Workshops (ICCVW)}, pages 4469--4478, 2023.

\bibitem[Deng et~al.(2009)Deng, Dong, Socher, Li, Li, and Fei-Fei]{deng2009imagenet}
Jia Deng, Wei Dong, Richard Socher, Li-Jia Li, Kai Li, and Li Fei-Fei.
\newblock {ImageNet}: A large-scale hierarchical image database.
\newblock In \emph{Proceedings of the IEEE Conference on Computer Vision and Pattern Recognition (CVPR)}, pages 248--255, 2009.

\bibitem[Fu et~al.(2024)Fu, Hu, Li, Feng, Wang, Lin, Roth, Smith, Ma, and Krishna]{fu2024blink}
Xingyu Fu, Yushi Hu, Bangzheng Li, Yu Feng, Haoyu Wang, Xudong Lin, Dan Roth, Noah~A. Smith, Wei-Chiu Ma, and Ranjay Krishna.
\newblock {BLINK}: Multimodal large language models can see but not perceive.
\newblock In \emph{Computer Vision -- ECCV 2024}, pages 148--166, 2024.

\bibitem[Gurari et~al.(2018)Gurari, Li, Stangl, Guo, Lin, Grauman, Luo, and Bigham]{gurari2018vizwiz}
Danna Gurari, Qing Li, Abigale~J. Stangl, Anhong Guo, Chi Lin, Kristen Grauman, Jiebo Luo, and Jeffrey~P. Bigham.
\newblock {VizWiz} grand challenge: Answering visual questions from blind people.
\newblock In \emph{Proceedings of the IEEE/CVF Conference on Computer Vision and Pattern Recognition (CVPR)}, pages 3608--3617, 2018.

\bibitem[Hudson and Manning(2019)]{hudson2019gqa}
Drew~A. Hudson and Christopher~D. Manning.
\newblock {GQA}: A new dataset for real-world visual reasoning and compositional question answering.
\newblock In \emph{Proceedings of the IEEE/CVF Conference on Computer Vision and Pattern Recognition (CVPR)}, pages 6700--6709, 2019.

\bibitem[Kuznetsova et~al.(2020)Kuznetsova, Rom, Alldrin, Uijlings, Krasin, Pont-Tuset, Kamali, Popov, Malloci, Kolesnikov, Duerig, and Ferrari]{kuznetsova2020openimages}
Alina Kuznetsova, Hassan Rom, Neil Alldrin, Jasper Uijlings, Ivan Krasin, Jordi Pont-Tuset, Shahab Kamali, Stefan Popov, Matteo Malloci, Alexander Kolesnikov, Tom Duerig, and Vittorio Ferrari.
\newblock The open images dataset {V4}: Unified image classification, object detection, and visual relationship detection at scale.
\newblock \emph{International Journal of Computer Vision}, 128\penalty0 (7):\penalty0 1956--1981, 2020.

\bibitem[Li et~al.(2024)Li, Lin, Peng, Nyandwi, Jiang, Ma, Khanuja, Krishna, Neubig, and Ramanan]{li2024naturalbench}
Baiqi Li, Zhiqiu Lin, Wenxuan Peng, Jean de~Dieu Nyandwi, Daniel Jiang, Zixian Ma, Simran Khanuja, Ranjay Krishna, Graham Neubig, and Deva Ramanan.
\newblock {NaturalBench}: Evaluating vision-language models on natural adversarial samples.
\newblock In \emph{Advances in Neural Information Processing Systems (NeurIPS)}, pages 17044--17068, 2024.

\bibitem[Li et~al.(2023)Li, Li, Savarese, and Hoi]{li2023blip2}
Junnan Li, Dongxu Li, Silvio Savarese, and Steven Hoi.
\newblock {BLIP-2}: Bootstrapping language-image pre-training with frozen image encoders and large language models, 2023.
\newblock arXiv:2301.12597.

\bibitem[Liu et~al.(2023)Liu, Li, Wu, and Lee]{liu2023visual}
Haotian Liu, Chunyuan Li, Qingyang Wu, and Yong~Jae Lee.
\newblock Visual instruction tuning.
\newblock In \emph{Advances in Neural Information Processing Systems (NeurIPS)}, 2023.

\bibitem[Liu et~al.(2024)Liu, Li, Li, and Lee]{liu2024improved}
Haotian Liu, Chunyuan Li, Yuheng Li, and Yong~Jae Lee.
\newblock Improved baselines with visual instruction tuning.
\newblock In \emph{Proceedings of the IEEE/CVF Conference on Computer Vision and Pattern Recognition (CVPR)}, 2024.

\bibitem[Parikh et~al.(2026)Parikh, Fein-Ashley, Kannan, and Prasanna]{parikh2026latent}
Dhruv Parikh, Jacob Fein-Ashley, Rajgopal Kannan, and Viktor Prasanna.
\newblock Latent denoising improves visual alignment in large multimodal models, 2026.
\newblock arXiv:2604.21343.

\bibitem[Sim{\'e}oni et~al.(2025)Sim{\'e}oni, Vo, Seitzer, Baldassarre, et~al.]{simeoni2025dinov3}
Oriane Sim{\'e}oni, Huy~V. Vo, Maximilian Seitzer, Federico Baldassarre, et~al.
\newblock {DINOv3}, 2025.
\newblock arXiv:2508.10104.

\bibitem[Song et~al.(2026)Song, Li, Ma, Tang, Wang, Zhang, Yang, Xiao, Liu, Zhang, Zhang, Zhang, Xu, Jiang, Han, Sun, Zhang, Huang, and Duan]{song2026joyai}
Lin Song, Wenbo Li, Guoqing Ma, Wei Tang, Bo Wang, Yuan Zhang, Yijun Yang, Yicheng Xiao, Jianhui Liu, Yanbing Zhang, Guohui Zhang, Wenhu Zhang, Hang Xu, Nan Jiang, Xin Han, Haoze Sun, Maoquan Zhang, Haoyang Huang, and Nan Duan.
\newblock {JoyAI-Image}: Awaking spatial intelligence in unified multimodal understanding and generation, 2026.
\newblock arXiv:2605.04128.

\bibitem[Tang et~al.(2026)Tang, Chen, Zhai, Wei, Liu, Zhao, Wu, Xiao, and Chen]{tang2026robustu1}
Jiaqi Tang, Jianmin Chen, Youyang Zhai, Wei Wei, Runtao Liu, Mengjie Zhao, Xiangyu Wu, Qingfa Xiao, and Qifeng Chen.
\newblock {Robust-U1}: Can {MLLMs} self-recover corrupted visual content for robust understanding?, 2026.
\newblock arXiv:2606.08063.

\bibitem[Tsimpoukelli et~al.(2021)Tsimpoukelli, Menick, Cabi, Eslami, Vinyals, and Hill]{tsimpoukelli2021frozen}
Maria Tsimpoukelli, Jacob Menick, Serkan Cabi, S.~M.~Ali Eslami, Oriol Vinyals, and Felix Hill.
\newblock Multimodal few-shot learning with frozen language models.
\newblock In \emph{Advances in Neural Information Processing Systems (NeurIPS)}, pages 200--212, 2021.

\bibitem[Wu et~al.(2024)Wu, Zhang, Zhang, Chen, Liao, Wang, Xu, Li, Hou, Zhai, Xue, Sun, Yan, and Lin]{wu2024qinstruct}
Haoning Wu, Zicheng Zhang, Erli Zhang, Chaofeng Chen, Liang Liao, Annan Wang, Kaixin Xu, Chunyi Li, Jingwen Hou, Guangtao Zhai, Geng Xue, Wenxiu Sun, Qiong Yan, and Weisi Lin.
\newblock {Q-Instruct}: Improving low-level visual abilities for multi-modality foundation models.
\newblock In \emph{Proceedings of the IEEE/CVF Conference on Computer Vision and Pattern Recognition (CVPR)}, pages 25490--25500, 2024.

\bibitem[Yang et~al.(2024)Yang, Luo, Li, Deng, Guibas, Krishnan, Weinberger, Tian, and Wang]{yang2024dvt}
Jiawei Yang, Katie~Z. Luo, Jiefeng Li, Congyue Deng, Leonidas~J. Guibas, Dilip Krishnan, Kilian~Q. Weinberger, Yonglong Tian, and Yue Wang.
\newblock Denoising vision transformers, 2024.
\newblock arXiv:2401.02957.

\bibitem[Yao et~al.(2024)Yao, Li, Ren, Wang, Liu, Sun, and Hou]{yao2024deco}
Linli Yao, Lei Li, Shuhuai Ren, Lean Wang, Yuanxin Liu, Xu Sun, and Lu Hou.
\newblock {DeCo}: Decoupling token compression from semantic abstraction in multimodal large language models, 2024.
\newblock arXiv:2405.20985.

\end{thebibliography}
